\documentclass[sn-basic]{sn-jnl}

\usepackage{graphicx}%
\usepackage{multirow}%
\usepackage{amsmath,amssymb,amsfonts}%
\usepackage{amsthm}%
\usepackage{mathrsfs}%
\usepackage[title]{appendix}%
\usepackage{xcolor}%
\usepackage{textcomp}%
\usepackage{manyfoot}%
\usepackage{booktabs}%
\usepackage{algorithm}%
\usepackage{algorithmicx}%
\usepackage{algpseudocode}%
\usepackage{listings}%
\usepackage[inline]{enumitem}%
\usepackage{caption}%
\usepackage{subcaption}%
\usepackage{graphicx}%
\usepackage{array,longtable}%
\usepackage{soul}%
\usepackage{pgfplots}%
\usepackage{makecell}%
\usepackage{tabularray}%
\usepackage{xurl}%
\usepackage{multirow}%
\usepackage[edges]{forest}%

\begin{document}

\title[Multimodal Explainable Artificial Intelligence: A Comprehensive Review of Methodological Advances and Future Research Directions]{Multimodal Explainable Artificial Intelligence: A Comprehensive Review of Methodological Advances and Future Research Directions}

\author[1]{\fnm{Nikolaos} \sur{Rodis}}\email{rodisnick@hua.gr}
\author[1]{\fnm{Christos} \sur{Sardianos}}\email{sardianos@hua.gr}
\author[2,3]{\fnm{Panagiotis} \sur{Radoglou-Grammatikis}}\email{pradoglou@uowm.gr, pradoglou@k3y.bg}
\author[2]{\fnm{Panagiotis} \sur{Sarigiannidis}}\email{psarigiannidis@uowm.gr}
\author[1]{\fnm{Iraklis} \sur{Varlamis}}\email{varlamis@hua.gr}
\author*[1]{\fnm{Georgios Th.} \sur{Papadopoulos}}\email{g.th.papadopoulos@hua.gr}

\affil[1]{\orgdiv{Department of Informatics and Telematics}, \orgname{Harokopio University of Athens}, \orgaddress{\street{Thiseos 70}, \city{Athens}, \postcode{GR 17676}, \state{Attiki}, \country{Greece}}}

\affil[2]{\orgdiv{Department of Electrical and Computer Engineering}, \orgname{University of Western Macedonia}, \orgaddress{\street{Active Urban Planning Zone}, \city{Kozani}, \postcode{GR 50150}, \state{Kozani}, \country{Greece}}}

\affil[3]{\orgdiv{K3Y}, \orgaddress{\street{Studentski district, Vitosha quarter, bl. 9}, \city{Sofia}, \postcode{BG 1700}, \state{Sofia City Province}, \country{Bulgaria}}}

\abstract{Despite the fact that Artificial Intelligence (AI) has boosted the achievement of remarkable results across numerous data analysis tasks, however, this is typically accompanied by a significant shortcoming in the exhibited transparency and trustworthiness of the developed systems. In order to address the latter challenge, the so-called eXplainable AI (XAI) research field has emerged, which aims, among others, at estimating meaningful explanations regarding the employed model’s reasoning process. The current study focuses on systematically analyzing the recent advances in the area of Multimodal XAI (MXAI), which comprises methods that involve multiple modalities in the primary prediction and explanation tasks. In particular, the relevant AI-boosted prediction tasks and publicly available datasets used for learning/evaluating explanations in multimodal scenarios are initially described. Subsequently, a systematic and comprehensive analysis of the MXAI methods of the literature is provided, taking into account the following key criteria: a) The number of the involved modalities (in the employed AI module), b) The processing stage at which explanations are generated, and c) The type of the adopted methodology (i.e. the actual mechanism and mathematical formalization) for producing explanations. Then, a thorough analysis of the metrics used for MXAI methods’ evaluation is performed. Finally, an extensive discussion regarding the current challenges and future research directions is provided.}

\keywords{Artificial intelligence, multimodal explainable artificial intelligence, deep learning, neural networks, explanation, evaluation}

\maketitle

\section{Introduction}	\label{intro}	

Over the last decade, humanity has witnessed unprecedented advancements in the field of Artificial Intelligence (AI), largely due to the emergence of the so-called Deep Learning (DL) paradigm that relies on the deployment of large-scale artificial neural networks and high-performing (GPU-enabled) computational infrastructures \cite{lecun2015deep}. The introduced algorithms have been adopted in numerous application areas, leading to ground-breaking solutions and tremendous performance improvements. For example, DL has revolutionized the nature of research in the fields of computer vision, Natural Language Processing (NLP), audio analysis, self-driving cars and robotics \cite{boukerche2021object, du2022elements, papCVPR, maruf2021survey, PapadopoulosAccess}, to name a few.

\begin{figure*} [t]
\centering
\begin{tabular}{cc}
\begin{tabular}{c}\begin{subfigure}[t]{0.35\textwidth}\includegraphics[width=\linewidth]{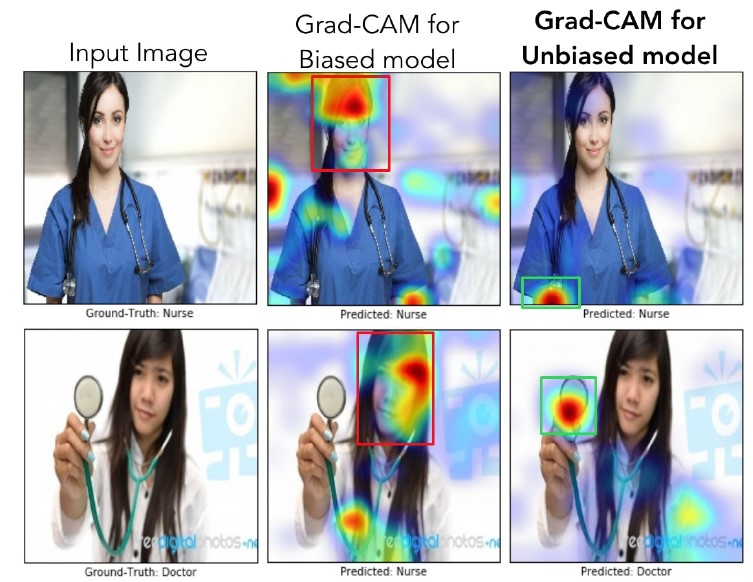}\caption{}\label{fig:bias_plot}\end{subfigure}\end{tabular}&
\begin{tabular}{c}\begin{subfigure}[t]{0.55\textwidth}\includegraphics[width=\linewidth]{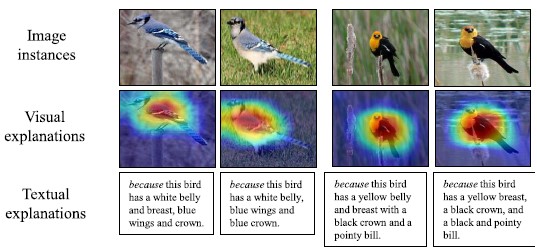}\caption{}\label{fig:MXAIexample}
\end{subfigure}\end{tabular}
\end{tabular}
\caption{Difference between unimodal and multimodal XAI: a) Unimodal explanation (saliency map) for image classification \cite{selvaraju2016grad2}, and b) Multimodal explanation (visual and text) zero-shot learning \cite{Liu2020}.}
\label{fig:examples}
\end{figure*}

Although AI-enabled solutions have consistently resulted into remarkable outcomes across various tasks to which they have been applied, this is, however, accompanied by a cost in the exhibited transparency and trustworthiness of the developed systems \cite{Arrieta2020}. In particular, it is typically very difficult to provide compact and precise explanations of the Neural Networks' (NNs) behavior and their eventual decision-making process. In order to address the latter challenge, the so-called eXplainable AI (XAI) research field has emerged, which aims, among others, at estimating meaningful explanations regarding the model's reasoning procedure \cite{Guidotti2018, Arrieta2020, adadi2018, minh2022explainable, di2023explainable, graziani2023global, kolajo2023human}. More specifically, how a prediction model works, how input data are used and what are the most critical/contributing features are some of the questions that XAI aims to answer. In this respect, XAI methods contribute towards making the models for the primary prediction task more transparent, while also significantly supporting the process of improving their performance. Indicative results of one of the most popular XAI methods are presented in Fig. \ref{fig:bias_plot}, where the Gradient-weighted Class Activation Mapping (Grad-CAM) approach is used for detecting and removing model bias in an image classification scenario \cite{Selvaraju2017}. 

While a large body of research works devoted to XAI has already been introduced, more recently the so-called Multimodal XAI (MXAI) approaches have been proposed, which naturally extend the fundamental principles and goals of unimodal XAI to the multimodal case (i.e. AI/DL-empowered methods that involve multiple types of modalities). The current work makes use of the term \textbf{\textit{modality}} as defined in \cite{baltruvsaitis2018multimodal}, which states that ``\textit{\textbf{Modality refers to the way in which something happens or is experienced and a research problem is characterized as multimodal when it includes multiple such modalities}}". In other words, the current work considers as MXAI methods those XAI approaches where: a) Multiple (two or more) modalities are used by the primary prediction model, b) Multiple (two or more) modalities are used for producing the primary models' behavior explanation, or c) The unimodal feature spaces of the primary model's input and the generated explanation are different (e.g. image/visual classification prediction (input) and textual explanation (output)). Therefore, MXAI refers to methods that overall involve multiple modalities in the primary and the explanation tasks. In this context, the most common modalities that are typically used in multimodal analytics schemes (and which in turn require multimodal analysis for explaining their behavior) include visual, audio, text, tabular, and graph data \cite{Liu2020, Kanehira2019, Hendricks2016, tsai-etal-2020-multimodal}.

Investigating the scope of MXAI in more detail, questions like ``How?", ``What?" and ``Why?" need to be addressed in the multimodal setting, i.e. taking into account the increased AI/ML model complexity and the respective data multimodality, which inevitably requires a different mathematical consideration than that in the conventional unimodal XAI case. For example, it is evident that the two modalities that are used in Fig. \ref{fig:MXAIexample} to explain a particular classification decision provide a more complete, justified and human-understandable explanation, compared to the respective one in Fig. \ref{fig:bias_plot}. Therefore, MXAI approaches can support explanations using multiple modalities that are complementary and cover more aspects of explainability. While aiming at developing robust MXAI schemes, the following key challenges (that also constitute clear advances and contradistinctions from the unimodal XAI case), need to be addressed (among others): 
\begin{itemize}
    \item Identification of the modality with the highest impact on the overall model's prediction;
    \item Discovery of the most salient features in each modality;
    \item Explanation of the employed modality fusion scheme itself, in order to understand the overall model's inference procedure and the cross-modal correlations that are learned;
    \item Detection of causal relations in the model's input-output data streams for generating explanations more easily conceivable by humans;
    \item Analysis of algorithms of increased complexity that are usually needed for handling multimodal data.
\end{itemize}
 
As clearly outlined by the above analysis, MXAI emerges as a rather challenging, yet promising, research field that has recently received increased attention, while its particular characteristics differentiate it significantly from the conventional unimodal XAI paradigm. In this context, this study aims to comprehensively investigate, summarize and analyze in depth all recent advances and current research trends in the field of MXAI. Specifically, the main contributions of this work are:
\begin{itemize}
    \item Formulation of a \textbf{comprehensive registry of AI/ML-boosted tasks} (i.e. specific application cases) where MXAI solutions have been applied until now, as well as a \textbf{thorough report of the relevant datasets} that have been used for learning/evaluating explanations in multimodal scenarios;
    \item \textbf{Systematic and comprehensive analysis/review of MXAI methods} that have been introduced so far, taking into account \textbf{three main criteria}: 
    \begin{itemize}
        \item the \textbf{number of involved modalities} (in the AI/ML module); 
        \item the \textbf{processing stage} at which explanations are generated;
        \item the \textbf{type of the adopted methodology} (i.e. the actual mechanism and mathematical formalization) for producing explanations;
    \end{itemize}
    \item Thorough examination of the \textbf{metrics used for MXAI methods' evaluation}; 
    \item Extensive discussion of \textbf{current challenges} and \textbf{future research directions} in the field.
\end{itemize}

It needs to be highlighted that the current work significantly extends the respective/previous review of the MXAI methods' landscape presented in \cite{Joshi_2021}, since the present study: a) Follows a more systematic approach for analyzing the literature grounded on three different criteria (namely the number of involved modalities, the explanation generation stage and the type of the adopted methodology), b) Focuses on investigating only MXAI methods', while maintaining not to revisit the relevant multimodal processing and unimodal XAI fields, and c) Provides a thorough analysis of the metrics used for quantitative MXAI evaluation.

The manuscript is organized as follows: Section \ref{tasks} discusses the various tasks and applications in which MXAI has been utilized, along with the respective publicly available datasets. Section \ref{methods} systematically presents the methods for generating explanations in multimodal scenarios. Section \ref{evaluation} details the metrics used for evaluating explanations in the multimodal setting. Finally, Section \ref{challenges} discusses the current challenges and future research directions in the field, while Section \ref{conclusion} concludes the paper.

\section{Prediction Tasks and Datasets Used in MXAI Scenarios} \label{tasks} 

MXAI approaches are employed for explaining the behavior of AI/ML prediction models that involve multiple types of modalities and/or producing explanations (of the primary AI model's behavior) using different or multiple modalities. In this respect, the most popular AI/ML tasks that exhibit the above characteristics and are shown to benefit from the generation of MXAI explanations are: Visual Question Answering (VQA) \cite{Park2018, patro2020, Goyal2016, Wu2019, alipour2020study, Lu2016HiCoAt, wu2020improving, anderson2018bottom, selvaraju2019taking, alipour2020impact, zhu2016visual7w, yang2016stacked, kazemi2017show, nagaraj-rao-etal-2021-first, ghosh2019generating, goyal2017making, li2018vqa, lu2022learn, patro2018differential, patro2019u, jung2021towards, Kim2017VisualEHadamard, trott2018interpretable, vedantam2019probabilistic, mascharka2018transparency, zhang2019interpretable}, visual captioning \cite{anderson2018bottom, selvaraju2019taking, Han2018, sun2020understanding, xu2015show, fang2019modularized, Petsiuk2018RISERI, ramanishka2017top, dong2017improving, ramanishka2017top}, visual common-sense reasoning (including also the visual dialog case extension) \cite{marasovic-etal-2020-natural, zellers2019recognition, Das_2017_CVPR}, recommendation systems \cite{chen2019personalized, lin2019explainableoutfit}, fine-grained visual classification and zero-shot learning \cite{Hassan2019, Liu2020, Hendricks2016, hendricks2018grounding, Selvaraju2018, Wickramanayake_Hsu_Lee_2019, hendricks2018counterfactuals, kanehira2019learning, barratt2017interpnet, xu2020model, Gulshad2020}, (human) activity recognition \cite{Kanehira2019, aakur2018inherently, zhuo2019explainable, Park2018}, emotion/sentiment recognition (including also hate speech detection and affect recognition) \cite{tsai-etal-2020-multimodal, zadeh2018multimodal, kumar2021towards, asokan2022interpretability, liu2022group, vijayaraghavan2021interpretable, wu2022interpretablecapsule, lin2019explainable}, candidate screening \cite{Kaya2017CVscreening}, self-driving cars \cite{Kim2018, Kim2020}, computer aided diagnosis \cite{wang2021, Lee2019CADmultimodal, lucieri2022exaid}, sleep range classification \cite{ellis2021novel}, and many more.

\begin{table*} [t]
        \footnotesize
        \caption{Datasets for MXAI learning and evaluation for a VQA task.}
        \label{tab:VQAdata}
        \centering    
        \begin{tblr}{
                colspec={X[1.5,l,m] X[1.8,l,m] X[1.5,l,m] X[5,j,m] }}
            \hline
            \textbf{Dataset (date)} & \textbf{Input space} & \textbf{Output space} & \textbf{Description} \\ \hline
             VQA-X (2018) \cite{Park2018}    & Image, textual questions & Textual answer, textual explanation & Extension of VQA 2.0 \cite{goyal2017making} (204,721 images, 1 million questions, 6.5 million of answers and 295,538 complementary QA pairs) with one textual explanation for each training QA pair, three for those in test/val sets and visual explanations collected from humans.  \\ \hline
            VQA-E (2018) \cite{li2018vqa}   & Image, textual questions &  Textual answer, textual explanation &  VQA 2.0 extension with explanations created from image/question/answer triplets. \\ \hline
             TextVQA-X (2021) \cite{nagaraj-rao-etal-2021-first}    & Image, textual questions & Textual answer, visual \& textual explanation & In addition to images (11,681 instances), questions (15,374 instances) and answers, it contains visual and textual explanations (67,055 instances each).  \\ \hline
            SCIENCEQA (2022) \cite{lu2022learn}  & Image and/or textual context, textual questions & Textual answer, lecture, text explanation & Multiple choice questions in various scientific fields annotated with lectures (17,798 instances) and textual explanations (19,202 instances). Some questions have image context (10,332 instances), some text context (10,220 instances) and some both (6,532 instances). \\ \hline
             CLEVR (2017) \cite{johnson2017clevr}    & Image, textual questions, symbolic program & Textual answer & Image (100,000 instances), questions (864,968 instances) and answers (849,980) for training and validation. Only images and questions for testing. Annotated scene graphs and symbolic program representations for images in training and validation sets.  \\\hline
             CLEVRCoGenT (2017) \cite{johnson2017clevr}  & Image, textual questions, symbolic program & Textual answer & Same with CLEVR, but the objects and their colors in images are captured under two different conditions.   \\\hline
             SHAPES (2016) \cite{andreas2016neural}  &  Image, textual questions, symbolic program & Textual answer & Yes/no answers and annotated programs for each question (244 unique questions for 15,616 images).  \\ \hline
            Visual genome (2017) \cite{krishna2017visual}   &Image, textual questions, scene graph, attributes & Textual answer & It contains images (108,077 instances) with region descriptions (4,3 million instances), objects (1,4 million instances), relations between objects (1,5 million instances), attributes (1,6 million instances), scenes (108,249 instances), region graphs (3,8 million instances) and question-answers (1,773,258 instances). \\\hline
             VQA-HAT (2017) \cite{Das2017HumanAttention} & Image, textual questions & Textual answer, attention map & Extension of the VQA dataset of \cite{antol2015vqa} (614,163 questions, 203,721 images and 10 answers for each train and validation question) with human attention maps for a subset of the training and the validation sets (58,475 and 1,374 instances, respectively).\\\hline
        \end{tblr}
\end{table*} 

Regarding the main datasets that have been introduced so far for developing and evaluating MXAI approaches, these are illustrated in Tables \ref{tab:VQAdata} and \ref{tab:datasets1}. In particular, the tables group the various datasets with respect to the corresponding application task (as discussed above), while they also include information about the type of the input/output modalities, date and a short description for each entry. It needs to be mentioned that Table \ref{tab:VQAdata} includes only datasets related to the VQA task, due to the increased popularity of this particular application case. On the other hand, it must be highlighted that additional relevant datasets are publicly available for each task (e.g. VQA); however, the ones indicated in Tables \ref{tab:VQAdata} and \ref{tab:datasets1} provide the framework for enabling MXAI evaluation (and not only the development of the related AI/ML module for the primary prediction task).

  \begin{table*} [t]
        \footnotesize
        \caption{Datasets for MXAI learning and evaluation for different primary prediction tasks.}
        \label{tab:datasets1}
        \centering
        \begin{tblr}{
                colspec={X[1.6,l,m] X[2.6,l,m] X[1.2,l,m] X[1.7,l,m] X[4.8,j,m]  }}
            \hline
            \textbf{Task} & \textbf{Dataset (date)} & \textbf{Input space} & \textbf{Output space} & \textbf{Description} \\ \hline
            \SetCell[r=4]{} Visual captioning & COCO-2017 (2017) \cite{lin2014microsoft}   & Image &  Caption, bbox, segmentation, labels & Object and stuff image segmentations from 80 and 91 categories, respectively, and 5 captions per image. \\ \hline
                        & COCO-2014 (2014) \cite{lin2014microsoft}  & Image & Caption, bbox, segmentation, labels  & Object and stuff image segmentations from 80 and 91 categories, respectively, and 5 captions per image.  \\\hline
            & Flirck30K entities (2015)  \cite{plummer2015flickr30k}  & Image & Caption, bbox & 244,000 co-reference chains for the Flickr30K captions \cite{Flirck30K}. \\ \hline
            & Visual genome (2017) \cite{krishna2017visual}   &Image, scene graph, attributes & Caption & See Table \ref{tab:VQAdata}\\\hline
            Zero-shot learning \& fine-grained classification &CUB (2011) \cite{wah2011caltech}  & Image, attributes & Labels, textual explanation, bbox & Bird images from 200 categories with 15 locations, 312 attributes, 1 bounding box and 5 sentences per image.\\\hline
            \SetCell[r=3]{}{Activity recognition}  & Olympic sports (2010) \cite{niebles2010modeling}   & Video &  Labels, bbox, attributes & Videos from 16 sport categories with attributes and bounding boxes. \\\hline
            & UCF101 (2012) \cite{UCF101}  & Video & Labels, bbox, attributes & Videos of 24 activities with attributes and bounding boxes. \\ \hline
             & ACT-X (2018) \cite{Park2018}  &Image& Labels, textual \& visual explanation & Images of 397 activities with descriptions, 3 textual explanations and visual explanations per image. \\ \hline
            Self-driving cars  & BDD-X (2018) \cite{Kim2018}  & Video & Actions, textual explanations & 77 hours of driving videos containing 3-4 actions each, annotated with descriptions and explanations. \\\hline
            Visual common-sense reasoning &  VCR (2019) \cite{zellers2019recognition}  &Image, multiple choice questions&Answer, textual rationale & 290,000 multiple choice image questions, including a rationale for the correct answer. \\\hline
            Breast mass diagnosis  & DDSM (1998) \cite{Heath1998}   &Image& Labels, textual explanation & 605 mass images with BI-RADS descriptions and mass locations. \\ \hline
            Outfit recommendations & ExpFashion (2019) \cite{lin2019explainableoutfit}  & Image & Recommend-ations, text explanations & 200,745 outfit images with at least 3 explanations for each.\\ \hline
        \end{tblr}
    \end{table*}

\section{Explainability in the multimodal setting} \label{methods}

This section investigates in depth the application of XAI methods under multimodal scenarios. In particular, Section \ref{MXAIdefinitions} provides the fundamental criteria that are considered for classifying the various MXAI approaches, as well as the resulting categories. Subsequently, Section \ref{conventionalXAI} investigates conventional unimodal XAI approaches that are directly, or after being extended in a straightforward way, applied to multimodal cases. Then, Sections \ref{MXAIcat1}, \ref{MXAIcat2} and \ref{Methodologies} analyze the MXAI literature in relation to the number of the involved modalities, the processing stage at which explanations are generated and the adopted methodology, respectively.

    \subsection{Criteria and Resulting Categories of MXAI Methods} \label{MXAIdefinitions}

\begin{figure*} [t]
\centering
\begin{tabular}{cc}
\begin{tabular}{c}\begin{subfigure}[t]{0.32\textwidth}\includegraphics[width=\linewidth]{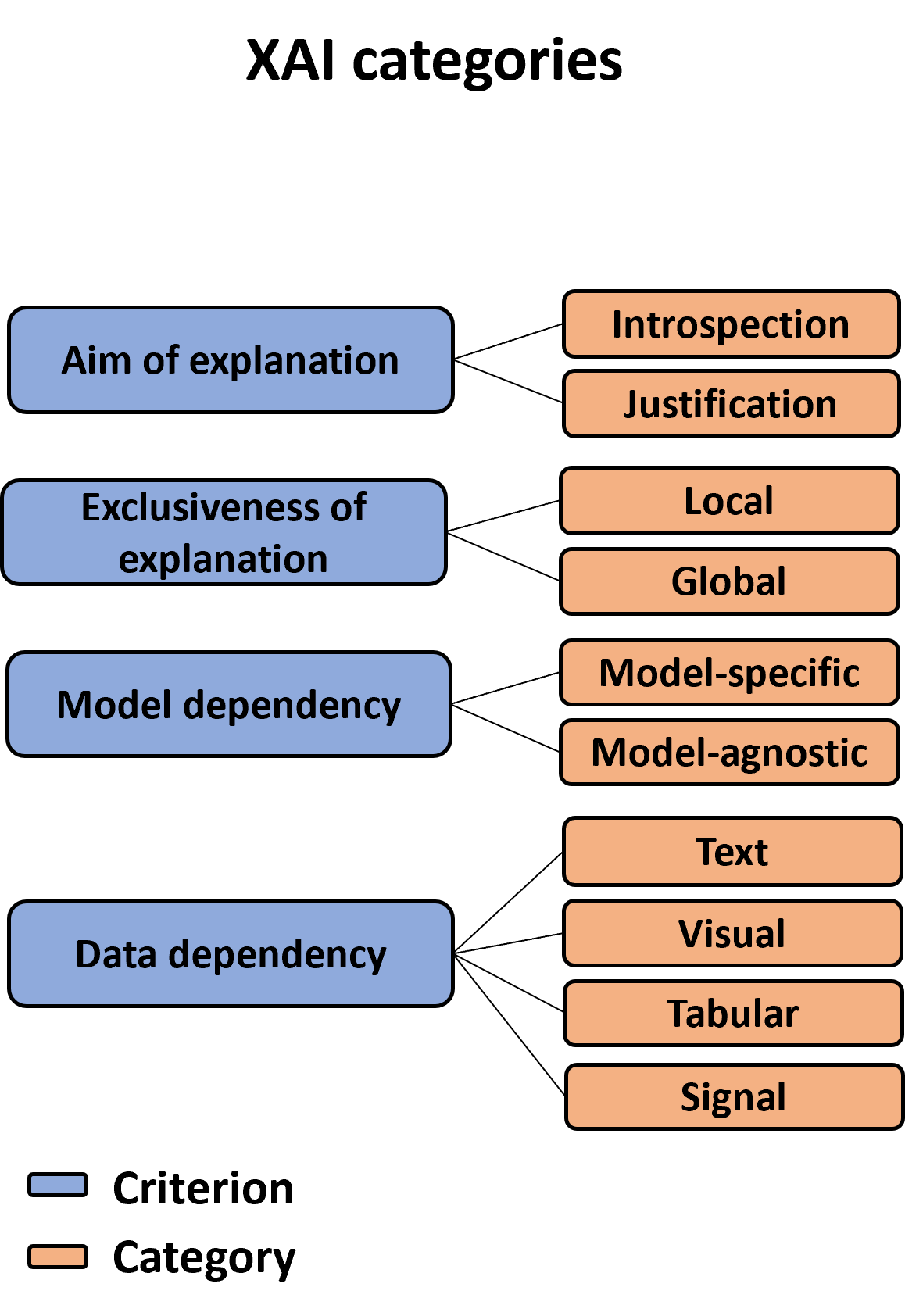}\caption{}\label{fig:GeneralXAI}\end{subfigure}\end{tabular}&
\begin{tabular}{c}\begin{subfigure}[t]{0.58\textwidth}\includegraphics[width=\linewidth]{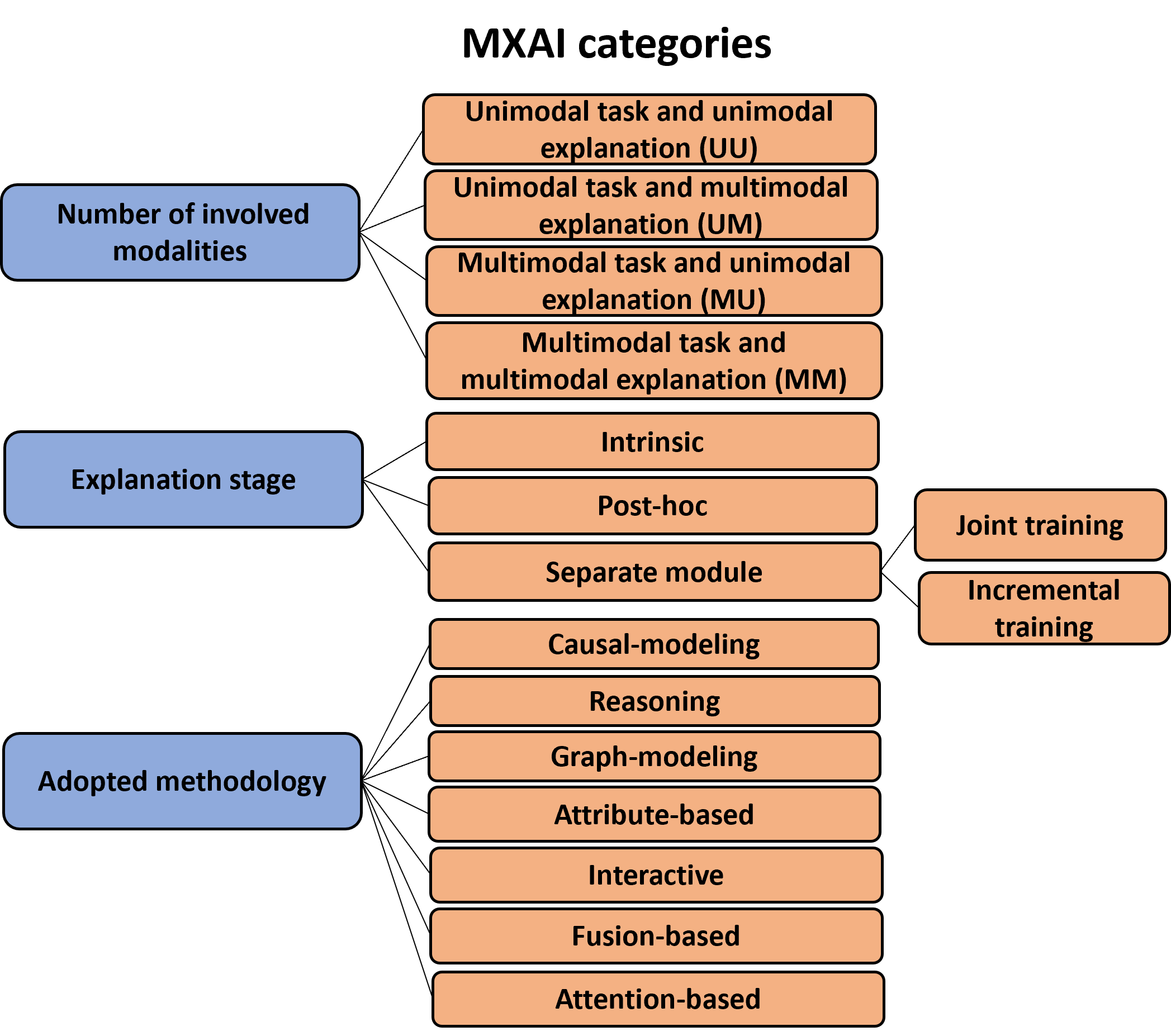}\caption{}\label{fig:MXAIgroups}\end{subfigure}\end{tabular}
\end{tabular}
\caption{Categorization of MXAI approaches: a) Main classes of conventional (unimodal) XAI methods that can also be adopted in MXAI analysis, b) Basic categories that can be considered only for MXAI approaches.}
\label{fig:taxonomy}
\end{figure*}

    Aiming at systematically analyzing the MXAI literature, an initial categorization can be made based on the classes of methods that have been widely adopted for the unimodal XAI case \cite{Hohman2019, adadi2018, Guidotti2018, Arrieta2020, Linardatos2020, Joshi_2021}. In particular, the following criteria, which are graphically illustrated in Fig. \ref{fig:GeneralXAI}, can be used:
    \begin{itemize}
    \item \textbf{Aim of explanation}: Depending on the scope of the explanation, MXAI methods can be considered as introspective \cite{Selvaraju2017} (i.e. focusing on the internal logic and behavior of the examined model) or justification-related \cite{Hendricks2016, Joshi_2021} (i.e. aiming at interpreting the model's predictions in a more human-friendly way).
    \item \textbf{Exclusiveness of explanation}: With respect to the extent of the validity of the generated explanation, MXAI methods can be either global \cite{tsai-etal-2020-multimodal} or local \cite{Lyu2022}. The former methods explain the overall model's behavior, while the latter focus on interpreting the produced decisions concerning specific data points.
    \item \textbf{Model dependency}: Concerning whether the internal structure and the architectural characteristics of the examined model are taken into account, MXAI approaches can be model-agnostic \cite{Lyu2022} (i.e. they can be applied to any type of ML model, regardless of its architecture or underlying algorithms) or model-specific \cite{Park2018} (i.e. they can be tailored to specific types/architectures of AI models).
    \item \textbf{Data dependency}: MXAI methods can be adapted so as to operate only for specific data types \cite{Linardatos2020}, e.g. visual, text, tabular, audio, etc., i.e. they can essentially be data type-specific approaches.
    \end{itemize}

\begin{figure*} [t]
\centering

\begin{tabular}{cc}
\begin{tabular}{c}\begin{subfigure}[t]{0.42\textwidth}\includegraphics[width=\linewidth]{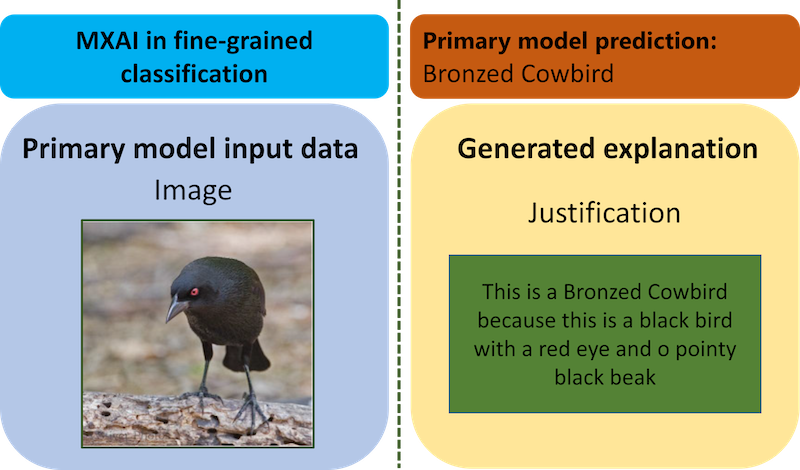}\caption{}\label{fig:uu}\end{subfigure}\end{tabular}&
\begin{tabular}{c}\begin{subfigure}[t]{0.48\textwidth}\includegraphics[width=\linewidth]{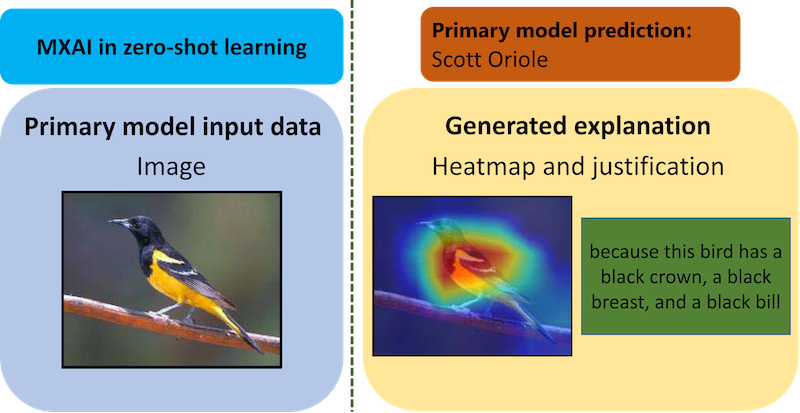}\caption{}\label{fig:um}\end{subfigure}\end{tabular}\\
\begin{tabular}{c}\begin{subfigure}[t]{0.42\textwidth}\includegraphics[width=\linewidth]{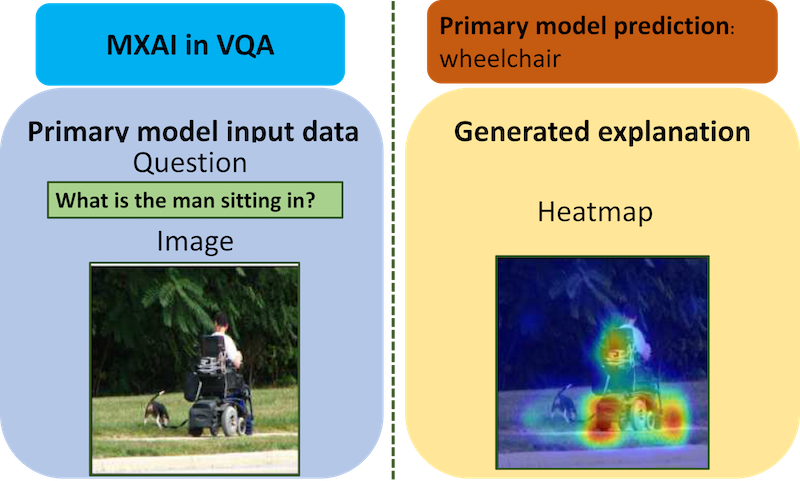}\caption{}\label{fig:mu}\end{subfigure}\end{tabular}&
\begin{tabular}{c}\begin{subfigure}[t]{0.48\textwidth}\includegraphics[width=\linewidth]{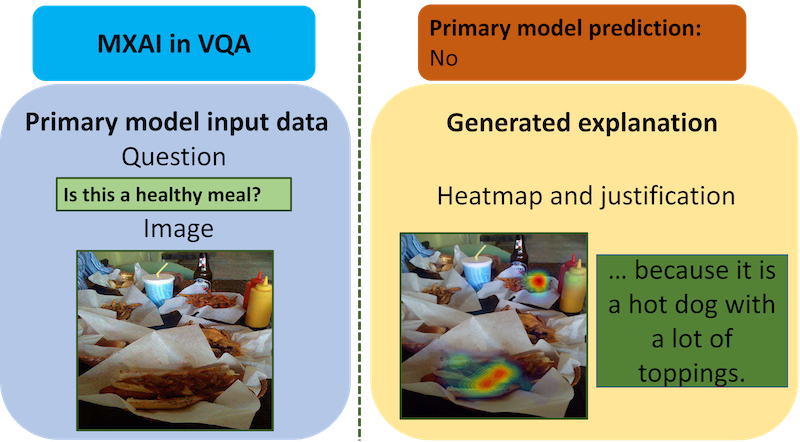}\caption{}\label{fig:mm}\end{subfigure}\end{tabular}\\
\end{tabular}
\caption{MXAI categories with respect to the number of the involved modalities in the primary prediction model input and the generated explanation: a) Unimodal task and unimodal explanation (UU) \cite{Hendricks2016}, b) Unimodal task and multimodal explanation (UM) \cite{Liu2020}, c) Multimodal task and unimodal explanation (MU) \cite{patro2018differential}, and d) Multimodal task and multimodal explanation (MM) \cite{Park2018}.}
\label{fig:MXAIcategories}
\end{figure*}

A critical characteristic of any MXAI method is the number of modalities that are being involved, considering both those that relate to the primary prediction task and those associated with the generated explanation. To this end, the following four main categories can be identified (Fig. \ref{fig:MXAIgroups}), while representative examples of each category are also graphically illustrated in Fig. \ref{fig:MXAIcategories}:
    \begin{itemize}
        \item \textbf{Unimodal task and unimodal explanation (UU)}: Comprises methods that for a unimodal primary prediction task estimate an explanation using a single, but different, modality. Fig. \ref{fig:uu} illustrates an example of an image classification algorithm (primary task), associated with a textual description/justification of the produced prediction (generated explanation) \cite{Hendricks2016}.
        \item \textbf{Unimodal task and multimodal explanation (UM)}: Includes approaches that for a unimodal primary task produce an explanation consisting of at least two modalities. Fig. \ref{fig:um} depicts an example of an image classification model (primary task), accompanied by a saliency map and a textual rationale description of the estimated prediction (generated explanation) \cite{Liu2020}. 
        \item \textbf{Multimodal task and unimodal explanation (MU)}: Comprises methods that for a multimodal primary task employing at least two modalities produce a unimodal explanation. Fig. \ref{fig:mu} illustrates an example of a VQA approach that receives as input an image and an associated question in textual format (primary task), while an image saliency map is estimated for interpreting the model's output (generated explanation) \cite{patro2018differential}.
        \item \textbf{Multimodal task and multimodal explanation (MM)}: Includes approaches that for a multimodal primary task (using at least two modalities) estimate also a multimodal explanation. Fig. \ref{fig:mm} illustrates an example of a VQA method, which is associated with both a saliency map and a textual rationale justification (generated explanation) \cite{Park2018}.
    \end{itemize}

    The development/processing stage (with respect to the model for the primary task) at which explanations are produced significantly differentiates the design of the corresponding MXAI approaches. Using the latter as a criterion, the following three main categories can be defined (in Fig. \ref{fig:MXAIgroups}):
    \begin{itemize}
        \item \textbf{Intrinsic}: Comprises methods that produce explanations by analyzing the internal structure and the parameters of the model that has been developed for the primary task \cite{Joshi_2021, adadi2018, asokan2022interpretability}.
        \item \textbf{Post-hoc}: Corresponds to those methods that do not investigate the internal architecture of the original model (developed for the primary task), i.e. approaches that are solely based on the analysis of the primary model's output \cite{sun2020understanding, ramanishka2017top}.
        \item \textbf{Separate module}: Incorporates methods that develop a distinct model (i.e. different from the one deployed for the primary task), in order to generate the required explanations \cite{Park2018, Lyu2022}. Depending on the actual phase during which the (new) explanation module is constructed, the approaches under this category can be further divided into:
        \begin{itemize}
            \item Joint training, where the explanation module is trained along with the model used for the primary task \cite{Hendricks2016, barratt2017interpnet};
            \item Incremental training, where the explanation module is constructed after the model for the primary task has been developed \cite{Kanehira2019, Wu2019}.
        \end{itemize}
    \end{itemize}
    
    MXAI methods can also be classified based on commonly met methodologies (i.e. mathematical formalizations and mechanisms) that constitute fundamental building blocks of their processing pipeline. In particular, some of the proposed methodologies (e.g. causal-modeling \cite{Kanehira2019}, reasoning \cite{Hendricks2016}, graph-modeling \cite{ghosh2019generating} and attribute-based \cite{Liu2020}) highly depend on the utilized data, while others (e.g. interactive \cite{alipour2020study}, fusion \cite{zadeh2018multimodal} and attention-based \cite{Lu2016HiCoAt}) are mostly related to the architecture of the examined model for the primary prediction task.

    Table \ref{tab:methodstable} summarizes the main MXAI approaches of the literature, which are hierarchically organized based on the number of the involved modalities, the explanation stage and the adopted methodology.

    \begin{table*} [h]
    \caption{Categorization of MXAI methods based on the number of involved modalities, explanation stage and adopted methodology.}
    \label{tab:methodstable}
    \footnotesize
    \centering
        \begin{tblr}{colspec={X[0.6,c,m] X[0.8,l,m] X[4.6,l,m]}}\hline
        \textbf{Task \& explanation modalities}  & \textbf{Explanation stage} & \textbf{Methodology} \\ \hline
        \SetCell[r=2]{} UU &Joint training &Reasoning \cite{Hendricks2016, barratt2017interpnet}, graph/example-based \cite{lu2016visual}, graph/attribute/rule-based \cite{zhuo2019explainable} \\\hline
        &Incremental training&Counterfactual \cite{hendricks2018counterfactuals}\\ \hline
        \SetCell[r=3]{} UM &Post-hoc&Example-based \cite{Gulshad2020}, gradient/attribute-based \cite{xu2020model}, interactive/graph-based \cite{aakur2018inherently}\\ \hline
        &Joint training &Example/attribute-based \cite{Hassan2019}, gradient/attribute-based \cite{Liu2020}\\\hline
        &Incremental training&Gradient/attribute-based \cite{Liu2020}, counterfactual \cite{Kanehira2019}, attention \cite{Kim2018}, counterfactual/reasoning \cite{hendricks2018grounding}, attribute/example-based \cite{kanehira2019learning}\\\hline
        \SetCell[r=4]{} MU &Intrinsic&Concept-based \cite{tsai-etal-2020-multimodal}, fusion/graph-based \cite{zadeh2018multimodal}, 
        attention \cite{chen2019personalized,anderson2018bottom,alipour2020impact,zhu2016visual7w,yang2016stacked,kazemi2017show,xu2015show,vijayaraghavan2021interpretable,patro2018differential,patro2019u,MCB2016,zhang2019interpretable,mascharka2018transparency,xie2019visual}, attention/gradient-based \cite{selvaraju2019taking}, fusion/attention \cite{liu2022group}\\ \hline
        &Post-hoc&Clustering \cite{kumar2021towards}, ablation \cite{lin2019explainable, ellis2021novel}, graph-based \cite{ghosh2019generating}, concept-based \cite{asokan2022interpretability}, attention \cite{ramanishka2017top}\\\hline
        &Joint training &Reasoning \cite{wu2020improving, lu2022learn}, attention \cite{Han2018, trott2018interpretable}, example-based  \cite{goyal2017making}, interactive \cite{Das_2017_CVPR}
        , concept-based \cite{dong2017improving}, graph-based \cite{vedantam2019probabilistic}\\\hline
        &Incremental training&Reasoning \cite{li-etal-2018-tell}, graph-based/reasoning \cite{marasovic-etal-2020-natural}, attention/attribute-based \cite{fang2019modularized}\\\hline
        \SetCell[r=4]{} MM &Intrinsic&Attention \cite{Lu2016HiCoAt,cao2020behind}\\ \hline
        &Post-hoc&Gradient/occlusion \cite{Goyal2016}, gradient/attribute-based \cite{Selvaraju2018}, rule-based \cite{Kaya2017CVscreening}\\\hline
        &Joint training &Attention/reasoning \cite{Park2018,li2018vqa,Kim2020}, attention \cite{patro2020}, attention/interactive/graph-based \cite{alipour2020study}, graph-based \cite{nagaraj-rao-etal-2021-first}, reasoning \cite{zellers2019recognition}\\ \hline
        &Incremental training&Attention \cite{Park2018,Wu2019,Lee2019CADmultimodal}, gradient/attribute-based \cite{Wickramanayake_Hsu_Lee_2019} \\\hline
        \end{tblr}
\end{table*}

    \subsection{Unimodal XAI Methods Extended to Multimodal Scenarios} \label{conventionalXAI}

    Several conventional XAI approaches, despite having been originally developed for unimodal tasks, can be extended to the multimodal setting in a relatively straightforward way. In the following, key representative examples of such methods are discussed in more detail.

    \textbf{Disentangled Multimodal Explanations (DIME)}: It constitutes a local model-agnostic method that extends the fundamental idea of the Local  Interpretable Model-agnostic Explanations (LIME) \cite{ribeiro-etal-2016-trust} approach to multimodal scenarios. In particular, DIME \cite{Lyu2022} disentangles the examined model into unimodal contributions and multimodal interactions, assuming that the overall model is formed as the aggregation of them. More specifically, LIME is applied separately to each unimodal contribution and to the multimodal interaction of the resulting aggregation.
     
    \textbf{Gradient-weighted Class Activation Mapping (Grad-CAM)}: The fundamental conceptualization of the Class Activation Map (CAM) \cite{zhou2016learning} approach is applicable to any model comprised of convolutional and a Global Average Pooling (GAP) layers. In this respect, the Grad-CAM \cite{Selvaraju2017} approach follows a back-propagation gradient-based scheme, where salient points in the input data that lead to the achieved prediction are identified, regardless of the number of the involved modalities.
    
\begin{figure*} [t]
\centering

\begin{tabular}{cc}
\begin{tabular}{c}\begin{subfigure}[t]{0.45\textwidth}\includegraphics[width=\linewidth]{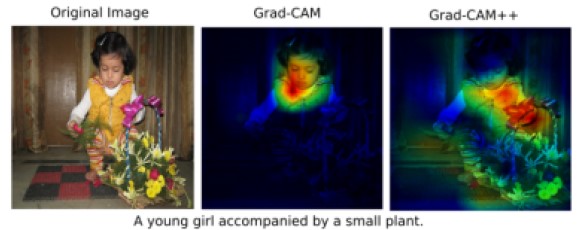}\caption{}\label{fig:camplus}\end{subfigure}\end{tabular}&
\begin{tabular}{c}\begin{subfigure}[t]{0.45\textwidth}\includegraphics[width=\linewidth]{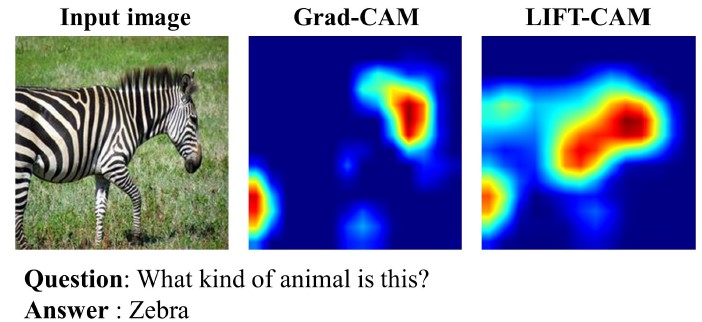}\caption{}\label{fig:LIFT-CAM}\end{subfigure}\end{tabular}\\
\begin{tabular}{c}\begin{subfigure}[t]{0.45\textwidth}\includegraphics[width=\linewidth]{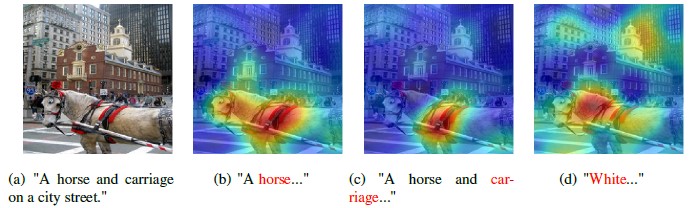}\caption{}\label{fig:rise}\end{subfigure}\end{tabular}&
\begin{tabular}{c}\begin{subfigure}[t]{0.45\textwidth}\includegraphics[width=\linewidth]{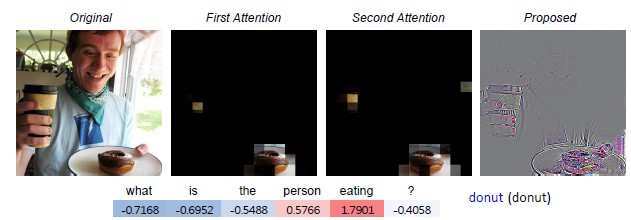}\caption{}\label{fig:Guided-Backpropagation}\end{subfigure}\end{tabular}\\
\end{tabular}
\caption{Conventional XAI methods' generated explanations for multimodal scenarios: a) Grad-CAM++ for image captioning \cite{Chattopadhay2018}, b) LIFT-CAM for a VQA task \cite{jung2021towards}, c) RISE for image captioning \cite{Petsiuk2018RISERI}, and d) Guided-backpropagation for a VQA task \cite{Kim2017VisualEHadamard}.}
\label{fig:XAIresults}
\end{figure*}

    \textbf{Grad-CAM++}: It is a modified version of the original Grad-CAM method, which uses a weighted average of the positive gradients of the target class. The updated method is able to locate multiple occurrences of the same type of objects in an image more accurately. Grad-CAM++ \cite{Chattopadhay2018} makes use of the same fundamental mechanism as Grad-CAM, which renders it possible to be applied to multimodal models as well, like those used for the tasks of VQA and image captioning (Fig. \ref{fig:camplus}).

    \textbf{LIFT-CAM}: The conventional Grad-CAM method \cite{Selvaraju2017} can be extended to the multimodal setting, as a combination of the SHapley Additive exPlanations (SHAP) \cite{Lundberg2017} and Deep Learning Important FeaTures (DeepLIFT) \cite{shrikumar2017DeepLIFT} approaches. In particular, LIFT-CAM \cite{jung2021towards} determines activation maps, making use of SHAP values; DeepLIFT is used in this case as an approximation to SHAP coefficients, due to their intractable nature (Fig. \ref{fig:LIFT-CAM}).
    
    \textbf{Combination of SHAP and Grad-CAM}: A combined approach of the conventional Grad-CAM and the SHAP methods is investigated in \cite{wang2021}, in order to provide explanations in a skin lesion diagnosis application. In particular, the SHAP method is used for handling tabular input data (e.g. age, gender, etc.), while Grad-CAM is employed for processing the visual input; hence, overall resulting in a multimodal explanation scheme.
    
    \textbf{Integrated Gradients (IG)}: The original method \cite{Sundararajan2017} aims to correlate the model's prediction results with the input data and to identify the most salient features, by estimating the path from the prediction output to the original input, using gradient-related information. Extending to the multimodal case, IG is applied in a VQA scenario in \cite{mudrakarta2018IGVQA}, in order to identify which words in the question are significant for producing the estimated answer.

    \textbf{Layerwise Relevance Propagation (LRP)}: The conventional LRP method relies on propagating the estimated prediction through the layers of the neural network \cite{Bach2015LRP}. Moving to the multimodal case, a variant of the LRP is used considering a convolutional neural network that receives as input three different sequences of MRI images, in order to detect characteristics of the brain that have an effect on its aging \cite{Hofmann2022LRPBrainAgeing}. In a similar way, multimodal LRP is also employed in \cite{sun2020understanding},  in order to identify both the important pixels in the image and the contribution of previously generated words in the produced caption.
     
    \textbf{Randomized Input Sampling for Explanation (RISE)}: The conventional RISE method \cite{Petsiuk2018RISERI} makes use of masked image inputs, in order to subsequently observe the resulting effect on the produced model's class prediction. The weighted aggregation of the masks and the respective prediction scores create the saliency map used for explanation. RISE's extension to the multimodal setting is straightforward. For example, saliency maps are estimated for each word in the produced textual description in an image captioning task \cite{Petsiuk2018RISERI} (Fig. \ref{fig:rise}).
    
    \textbf{Guided-backpropagation}: The original backpropagation method \cite{springenberg2015striving} aims at visualizing specific image features that have been detected from certain neurons, e.g. in an image analysis task. Its application to the multimodal setting is also relatively straightforward. For example, saliency maps are produced in a VQA task \cite{Kim2017VisualEHadamard}, where the Hadamard product of the visual and the textual features results in more accurate importance maps, compared to those resulted from using the respective attention weights \cite{Lu2016HiCoAt, nam2017dual} (Fig. \ref{fig:Guided-Backpropagation}). 

    \textbf{Concept Activation Vectors (CAV)}: The original method \cite{kim2018interpretability} aims at identifying specific semantic concepts, whose presence in the input data affects the model's prediction with respect to a given class. The latter relies on the use of differential/derivative estimations to assess how important is a certain pre-defined concept to a particular classification decision. In a multimodal scenario, CAV vectors can be used for explaining multimodal (audio, video and text data) emotion recognition, taking into account  concepts that are relevant to different types of emotions \cite{asokan2022interpretability}.

    \subsection{MXAI Categorization Based on the Number of the Involved Modalities}  \label{MXAIcat1} 
    
MXAI methods can be classified taking into account the different combinations of the number of modalities involved in the primary prediction task and the generated explanation, as discussed in Section \ref{MXAIdefinitions} and graphically illustrated in Fig. \ref{fig:MXAIcategories}.

        \subsubsection{Unimodal task and unimodal explanation (\textbf{UU})}

\begin{figure*} [t]
\centering
\begin{tabular}{cc}
\begin{tabular}{c}\begin{subfigure}[t]{0.45\textwidth}\includegraphics[width=\linewidth]{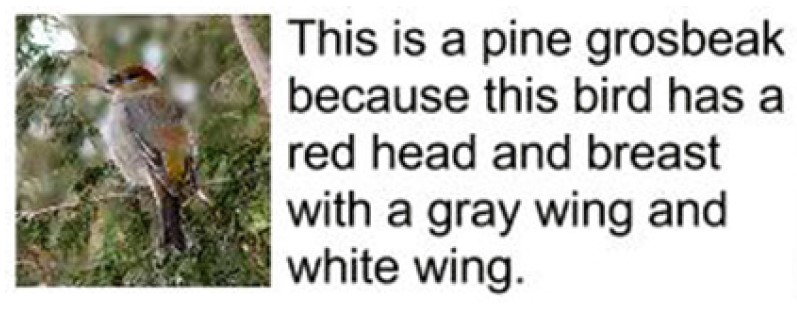}\caption{}\label{fig:uu1}\end{subfigure}\end{tabular}&
\begin{tabular}{c}\begin{subfigure}[t]{0.45\textwidth}\includegraphics[width=\linewidth]{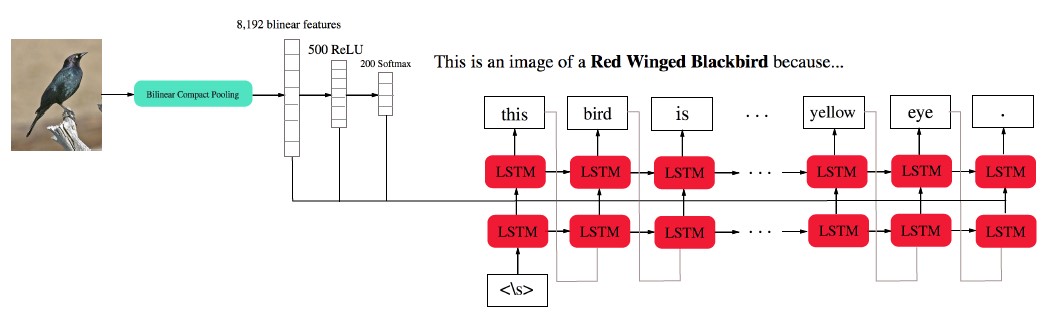}\caption{}\label{fig:uu3}\end{subfigure}\end{tabular}\\
\begin{tabular}{c}\begin{subfigure}[t]{0.45\textwidth}\includegraphics[width=\linewidth]{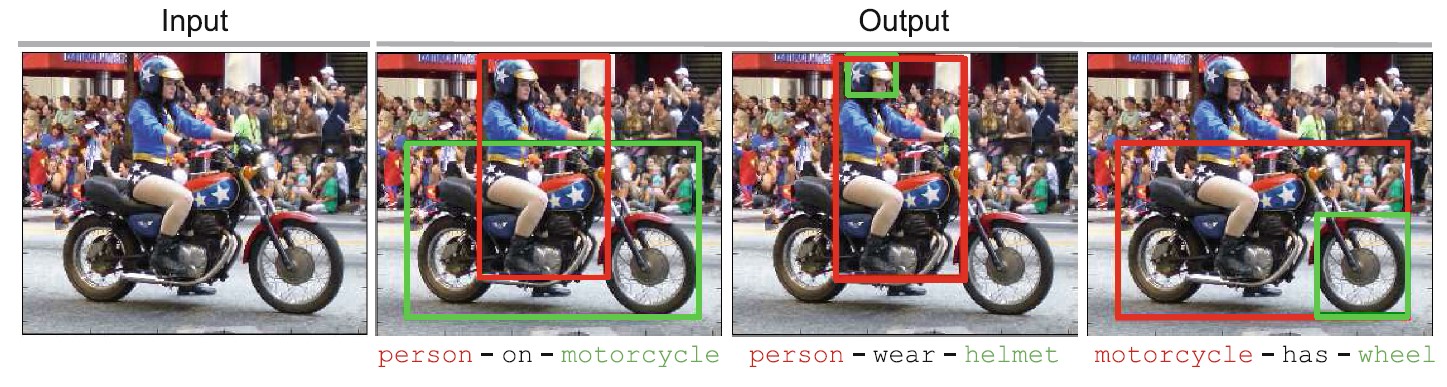}\caption{}\label{fig:uu2}\end{subfigure}\end{tabular}&
\begin{tabular}{c}\begin{subfigure}[t]{0.45\textwidth}\includegraphics[width=\linewidth]{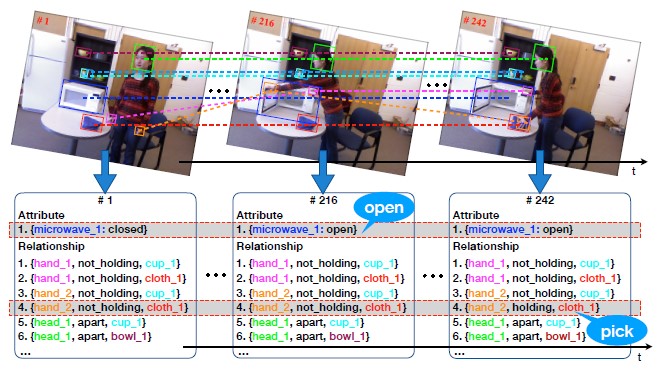}\caption{}\label{fig:uu4}\end{subfigure}\end{tabular}\\
\end{tabular}
\caption{UU MXAI methods' generated explanations: a) Textual \cite{Hendricks2016}, b) Textual \cite{barratt2017interpnet}, c) Object relationships \cite{lu2016visual}, and d) Scene graph \cite{zhuo2019explainable}.}
\label{fig:UUres}
\end{figure*}

        Methods belonging to this category often involve the visual modality as input for the primary prediction task, while the explanation can be either in textual format or in the form of a graph for efficiently representing correlations among entities. Fig. \ref{fig:UUres} illustrates indicative examples of representative literature approaches.

        \textbf{Textual explanations}: Explanations of models' predictions in textual form provide an efficient way for intuitively detailing the actual models' reasoning process. As an example, in a bird image classification task, the justifications of object classification decisions are generated based on the key discriminator factors of different bird species in \cite{Hendricks2016}; the method employs reinforcement learning techniques, while the generated explanations are class/prediction-specific, focusing on the explanation and not on a general-purpose image caption (Fig. \ref{fig:uu1}). Similarly, the activations of all neural network layers are concatenated in \cite{barratt2017interpnet}, targeting to provide a concrete explanation for the prediction of the bird class (Fig. \ref{fig:uu3}). \citet{hendricks2018counterfactuals} extend the latter idea to the case of estimating counterfactual explanations \cite{hendricks2018counterfactuals}, where missing characteristics/properties of the depicted objects are identified, aiming at providing a rationale regarding why a specific bird does not belong to a given category. In a different image-based task, where the goal is to train a recommender system to also provide meaningful explanations, user comments are treated as ground-truth explanations in \cite{lin2019explainableoutfit}, in order to justify the matching of top and bottom clothes (e.g. shirts and trousers).
        
        \textbf{Graph-based explanations}: Graph models provide an elegant way for formulating explanations, since they are inherently capable of representing multiple and diverse types of relationships among entities. Such an approach is particularly suitable for several application tasks requiring the justification of the prediction outcome based on the detected relationships, like in visual captioning, image classification and action recognition, to name a few. In \cite{lu2016visual}, an object-relation graph is created for connecting the objects detected in an image with various relationships (Fig. \ref{fig:uu2}), where three types of connections between objects in the examined image are supported. \citet{zhuo2019explainable} follow a graph-based approach in video-based action recognition, where for each input video frame the output consists of a scene graph depicting the relations among the objects inside the scenery (Fig. \ref{fig:uu4}).
        
        \subsubsection{Unimodal task and multimodal explanation (\textbf{UM})} 

        Explanations of AI models' behavior are usually in a unimodal form, which, however, can sometimes lead to incomplete representation and understanding of the model’s reasoning process. To this end, multimodal explanations can often be advantageous, since they provide additional/supplementary explanatory statements in different modalities. In the following, UM MXAI approaches, which are grouped with respect to the combination of modalities involved in the generated explanation, are discussed in detail.
        
\begin{figure*} [t]
\centering
\begin{tabular}{cc}
\begin{tabular}{c}\begin{subfigure}[t]{0.45\textwidth}\includegraphics[width=\linewidth]{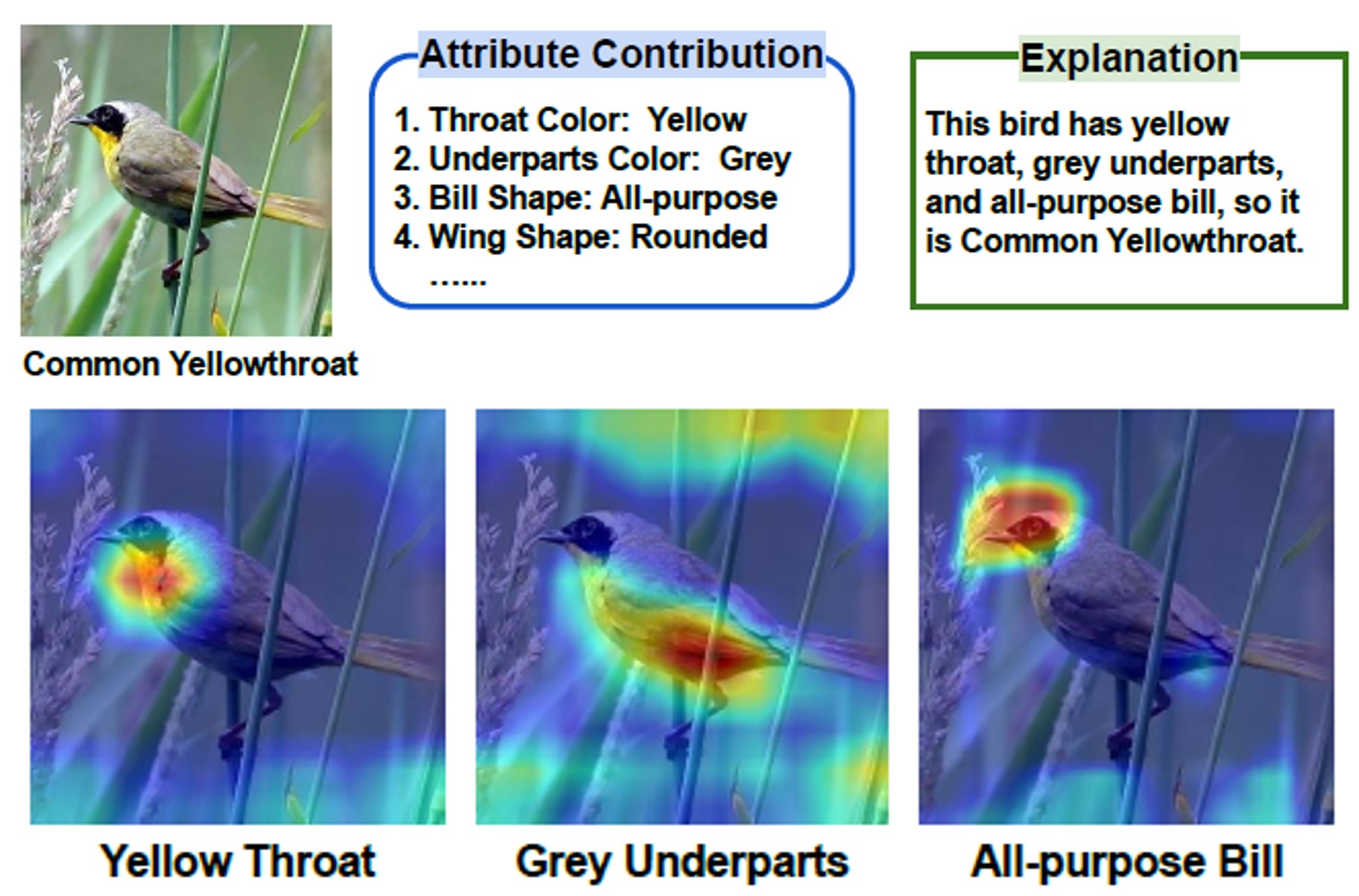}\caption{}\label{fig:um4}\end{subfigure}\end{tabular}&
\begin{tabular}{c}\begin{subfigure}[t]{0.45\textwidth}\includegraphics[width=\linewidth]{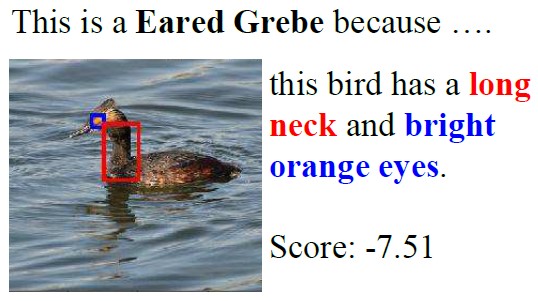}\caption{}\label{fig:um3}\end{subfigure}\end{tabular}\\
\begin{tabular}{c}\begin{subfigure}[t]{0.45\textwidth}\includegraphics[width=\linewidth]{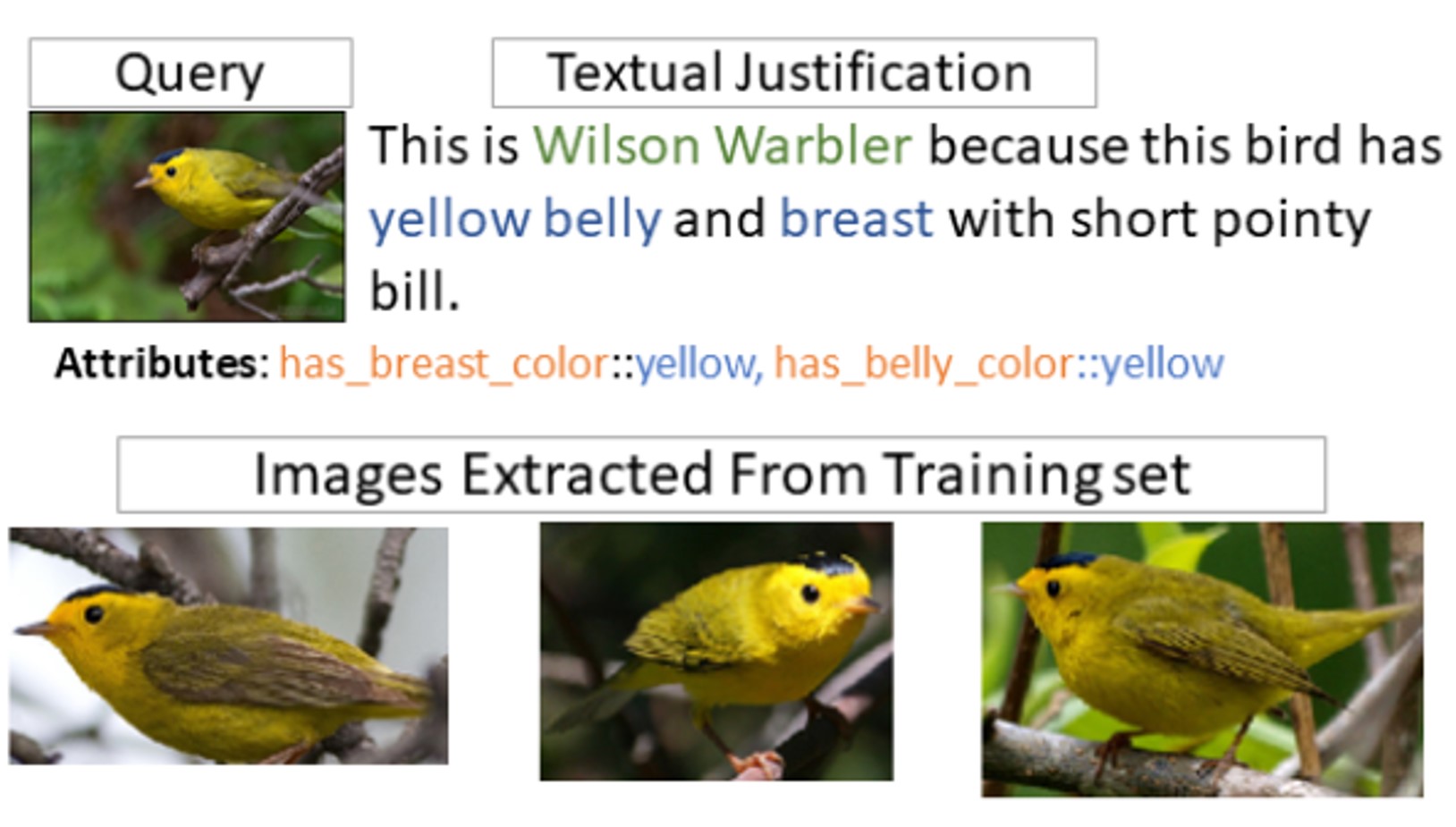}\caption{}\label{fig:um1}\end{subfigure}\end{tabular}&
\begin{tabular}{c}\begin{subfigure}[t]{0.45\textwidth}\includegraphics[width=\linewidth]{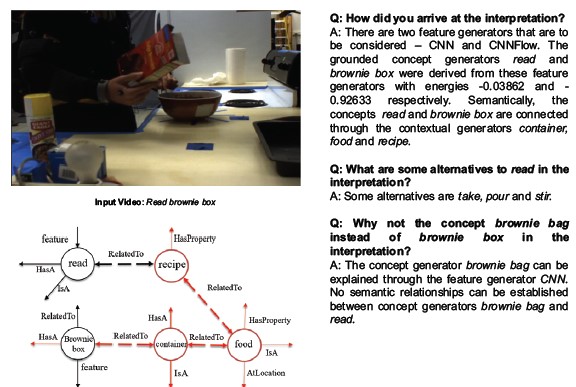}\caption{}\label{fig:um2}\end{subfigure}\end{tabular}\\
\end{tabular}
\caption{UM MXAI methods' generated explanations: a) Heatmap-text \cite{xu2020model}, b) Heatmap-text \cite{Kim2018}, c) Example-attribute \cite{Hassan2019}, and d) Graph-text \cite{aakur2018inherently}.}
\label{fig:UMres}
\end{figure*}

        \textbf{Heatmap-text explanation}: A reasoning approach is followed in \cite{xu2020model}, where the primary task attribute prediction is initially performed, while a combination of the prediction’s embeddings is subsequently used for producing the final classification. Then, using a back-propagation formalism, a score for each image attribute is obtained and the top-3 ones are subsequently utilized for forming a textual explanation, while Grad-CAM is also used for generating corresponding visual explanations (Fig. \ref{fig:um4}). Similarly, a reasoning-based MXAI method focusing on the activities of acceleration and course prediction in the scenario of a self-driving car is investigated in \cite{Kim2018}, where an attention map provides introspective inference behind the obtained prediction results, while text is also generated as a supporting means to justify them. Moreover, the Grad-CAM approach is employed within the zero-shot learning paradigm for estimating a CAM (i.e. heatmap) for each image attribute in \cite{Liu2020}, where a whole image CAM representation is obtained by merging the individual CAMs of all considered attributes to form a heatmap for the examined class, while a textual description is also generated using visual features, attributes and latent embeddings.

        \textbf{Image-text explanation}: Natural language explanations do not always provide a sufficient and complete justification of a model's particular decision; hence, grounding the produced explanations also on the visual modality is shown to be beneficial. In this respect, the methods of \cite{Hendricks2016} and \cite{hendricks2018grounding} generate textual explanations for the examined image, chunk them into phrases and estimate a score for each of them, indicating image relevance; the highest-scoring chunks are subsequently used for reinforcing the textual explanations (Fig. \ref{fig:um3}). In \cite{kanehira2019learning}, the textual part of the explanation is combined with complementary visual examples (e.g. images), which are associated with the generated textual justification; the estimated example can belong to the same or an opposing semantic class. Moreover, the textual justification is accompanied by an additional counterfactual part in \cite{Kanehira2019}, where action predictions in videos are justified using a combination of textual explanations and a bounding box of the predicted and the opposing classes.
        
        \textbf{Example-attribute explanation}: Aiming at increasing the expressiveness and the completeness of the generated explanations, estimated attributes (e.g. shape, color, etc.) can be combined with exemplary instances of the original input image data. In particular, \citet{Hassan2019} follows a visual search approach, apart from incorporating attributes for producing a justification of the model's prediction, in a visual classification task (Fig. \ref{fig:um1}). This visual search strategy relies on exploiting the convolutional features of the utilized NN model, in order to retrieve similar data instances that serve as additional justification information. Moreover, an adversarial approach is followed in \cite{Gulshad2020}, in order to estimate complementary/counter-examples, each associated with individual image attributes (e.g. bird bill shape, color, etc.).

        \textbf{Graph-text explanation}: The expressiveness of graphs in representing accurate explanations can be further reinforced, by combining them with an interaction mechanism. In particular, a video is provided as input to a network that outputs a graph, based on the detected objects and actions \cite{aakur2018inherently}. Apart from the explanation composed of the identified semantic concepts and their relations, the graph is utilized by an interactive question-answering agent that can answer inquiries regarding the graph structure and the video (Fig. \ref{fig:um2}); hence, producing additional textual explanations or alternative graph explanations.

        \subsubsection{Multimodal task and unimodal explanation (\textbf{MU})}

        MXAI techniques are particularly suitable for addressing explainability needs concerning multimodal prediction tasks, i.e. when the examined AI model receives as input data from multiple modalities. In the following, MU MXAI methods that produce unimodal explanations are discussed. For the sake of clarity, the presentation is organized according to the particular methodology/mechanism that is adopted for producing the explanations.
        
\begin{figure}[t]
\centering
\begin{tabular}{cc}
\begin{tabular}{c}\begin{subfigure}[t]{0.45\textwidth}\includegraphics[width=\linewidth]{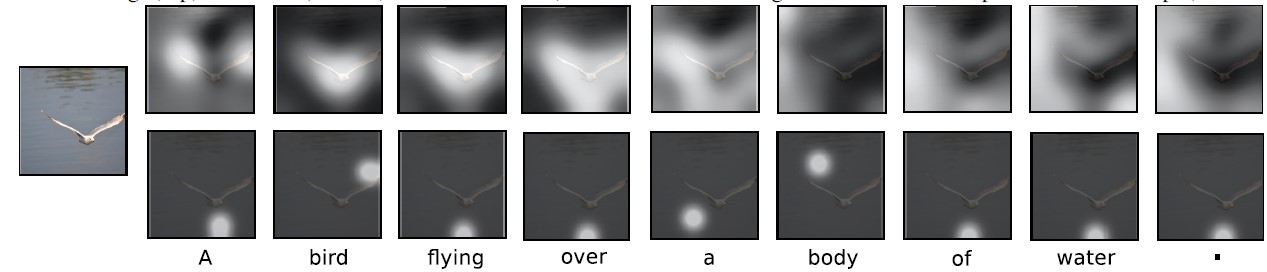}\caption{}\label{fig:mu5}\end{subfigure}\end{tabular}&
\begin{tabular}{c}\begin{subfigure}[t]{0.45\textwidth}\includegraphics[width=\linewidth]{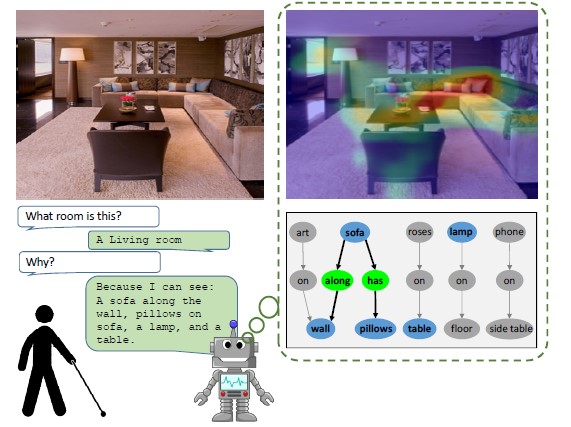}\caption{}\label{fig:mu8}\end{subfigure}\end{tabular}\\
\begin{tabular}{c}\begin{subfigure}[t]{0.45\textwidth}\includegraphics[width=\linewidth]{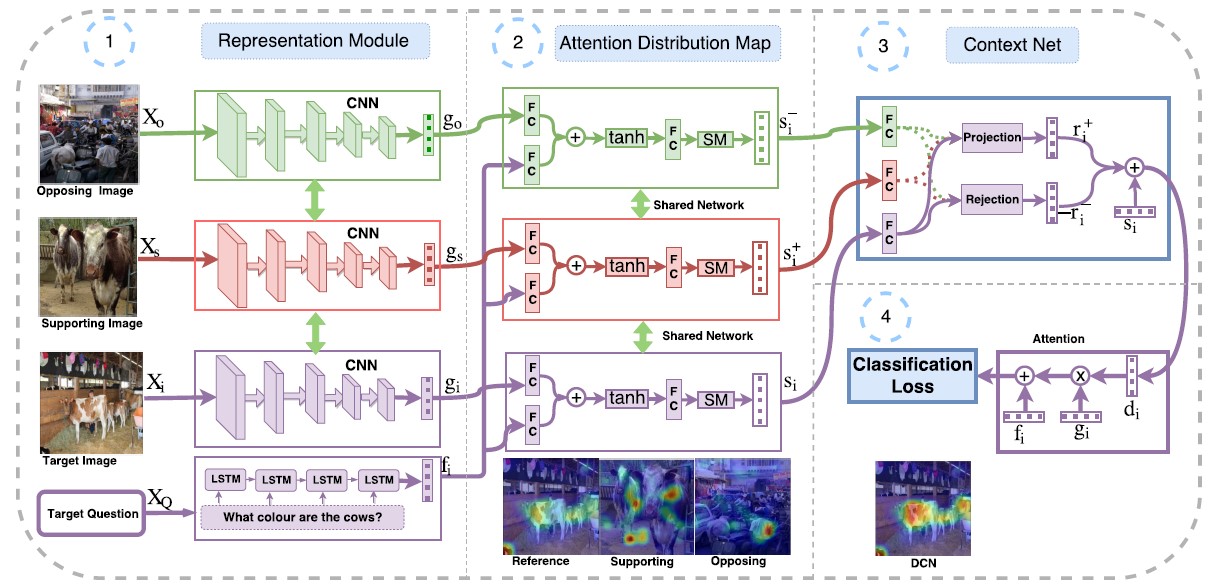}\caption{}\label{fig:mu4}\end{subfigure}\end{tabular}&
\begin{tabular}{c}\begin{subfigure}[t]{0.45\textwidth}\includegraphics[width=\linewidth]{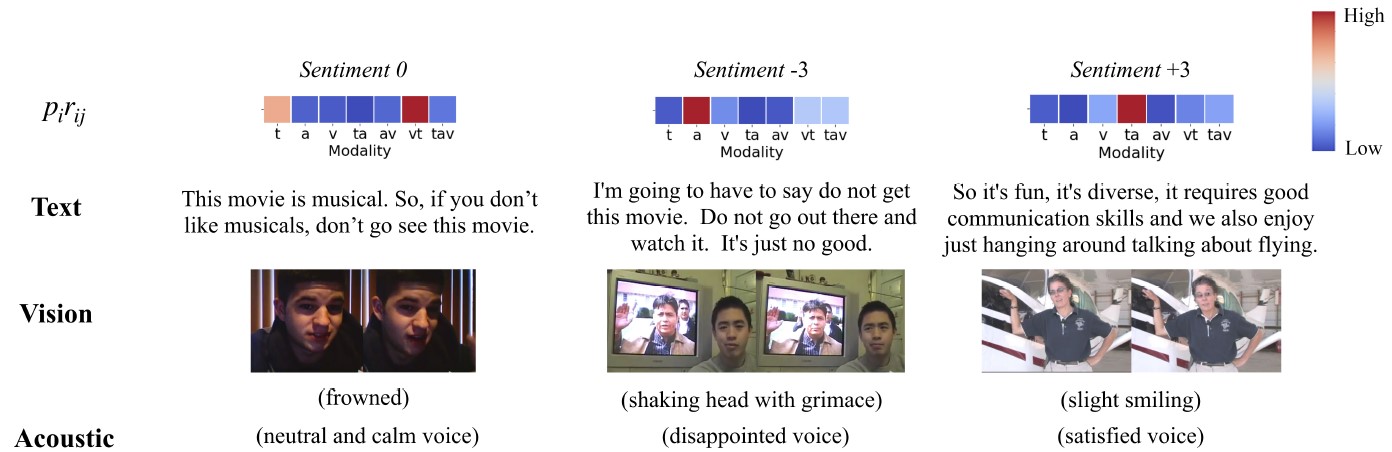}\caption{}\label{fig:mu1}\end{subfigure}\end{tabular}\\
\end{tabular}
\caption{MU MXAI methods' generated explanations: a) Attention map for image captioning \cite{xu2015show}, b) Textual explanation for a VQA task using graphs \cite{ghosh2019generating}, c) Example-based combined with attention mechanism for a VQA task \cite{patro2018differential}, and d) Concept-based mechanism for an interpretable emotion recognition task \cite{tsai-etal-2020-multimodal}.}
\label{fig:MUresA}
\end{figure}

        \textbf{Attention-based}: Attention schemes constitute the most widely used mechanism in MU MXAI methods, due to their inherent ability to adjust the analysis focus (during the explanation generation process) on the important parts of the input data (e.g. critical image regions and words). One of the most common primary tasks that attention mechanisms exhibit wide applicability is that of visual captioning. In particular, a stochastic and a deterministic attention mechanism are used in \cite{xu2015show}, in order to estimate heatmaps of the image regions that correspond to the predicted captions (Fig. \ref{fig:mu5}). Similarly, spatial and spatiotemporal saliency maps are computed for the estimated image and video captions in \cite{ramanishka2017top}, respectively. \citet{Han2018} compute relevance scores between the detected objects and the respective words in the estimated captions, aiming at explaining why certain words are generated. Moreover, attended regions in the visual medium are employed in captioning and VQA settings \cite{anderson2018bottom}, in order to interpret the produced captions or answers, respectively.

        Attention-based MU MXAI methods are also widely used in VQA applications, where the typical approach consists of estimating an attention map over the input image that highlights important regions that mostly contribute to the generated visual answer. Specifically, a pooling operator is used twice in the examined model in \cite{MCB2016}, where the attended regions are used for grounding the produced answer. Additionally, \citet{zhu2016visual7w} combine an attention map with a bounding box of the detected object for estimating a visual explanation. Multiple attention layers are used in \cite{yang2016stacked} and various attention distributions over the image are considered in \cite{kazemi2017show}, in order to localize the important image regions. Moreover, objects and regions of attention are used during the training phase of a visual grounding model in \cite{zhang2019interpretable}. \citet{alipour2020impact} introduce an explainable VQA system, in order to examine the impact of the explanations on the users in terms of model competency. Furthermore, important visual regions are estimated within a clothing recommendation system \cite{chen2019personalized}, analyzing both product images and customer reviews. 

        Among other application cases of MU MXAI methods that rely on the use of attention mechanisms, object detection and human-generated heatmaps are used by the so-called Human Importance-aware Network Tuning (HINT) \cite{selvaraju2019taking} approach, in order to improve performance in vision-language models, by leveraging gradient-based explanations. Additionally, the most contributing words in hate speech detection are estimated in \cite{vijayaraghavan2021interpretable}, using text, social and demographic features. Moreover, attention to both image and text is employed in \cite{xie2019visual}, in order to visualize regions of interest that correspond to a specific hypothesis in a visual entailment task.

        \textbf{Graph-modelling}: Graph models are suitable for producing unimodal explanations from a multimodal input feature space. In particular, bounding boxes are used to interpret the answer regarding the counting abilities (e.g. counting objects in images) of a model in \cite{trott2018interpretable}, while a text explanation is estimated using an image scene graph and an attention map in \cite{ghosh2019generating} (Fig. \ref{fig:mu8}). Additionally, \citet{vedantam2019probabilistic} follow a probabilistic approach to create an interpretable VQA model, using symbolic programs that can track the model’s reasoning. Symbolic programs are also used in \cite{mascharka2018transparency}, in order to generate visualizations of each step of the model’s inference procedure.
        
        \textbf{Reasoning}: An emerging line of research in the field of MXAI concerns the use of various reasoning formalisms, which often aim to mimic the human way of inference for generating improved explanations. In that respect, textual explanations provided by humans are used for creating a VQA model in \cite{wu2020improving}, which learns not only to estimate answers but also makes use of retrieved explanations to generate more accurate ones. Additionally, \citet{lu2022learn} produce textual explanations in response to scientific questions, making use of lectures (related to the subject in question), complementary to the use of relevant images and questions. Pre-trained transformer-based language models exploit objects and their relations to provide full-sentence answers and rationales in a VQA setting in \cite{marasovic-etal-2020-natural}. Moreover, a visual dialog approach is presented in \cite{Das_2017_CVPR}, where a conversation regarding the image at hand is produced, by answering consecutive questions about it.
    
        \textbf{Example-based}: An alternative approach, aiming at increasing the expressiveness of the generated explanations, consists of the use of exemplary data instances. In this context, counter-visual instances are used in a VQA scenario in \cite{goyal2017making}, targeting to both support the quality of the produced answers and to restrict the model's bias (with respect to the input questions), by collecting similar images for which for the same question a different answer is estimated. Additionally, images that support/oppose the produced answer in a VQA setting are used to create a map in \cite{patro2018differential}, which depicts the image regions that humans would have focused on (Fig. \ref{fig:mu4}).
        
        \textbf{Concept-based}: Explanations that are grounded on concept-based representations are often shown to be advantageous, regarding their interpretation by the human user. In this respect, semantic information, extracted using Latent Dirichlet Analysis (LDA), is considered for detecting activations that correspond to certain topics (e.g. people, dancing and eating, among others) in a video captioning setting in \cite{dong2017improving}; the explanation is eventually provided by a numerical metric that quantifies the correlation of a given neuron activation with a topic in a video frame. Additionally, \citet{tsai-etal-2020-multimodal} estimate local and global explanations for multimodal sentiment analysis using a routing approach to identify the importance of data (either characterizing individual modalities or cross-modal features), based on explainable hidden embeddings (Fig. \ref{fig:mu1}).
        
        \textbf{Fusion}: Fusion schemes are also shown to be beneficial in the explanation generation process. In particular, an explainable fusion scheme is proposed in \cite{zadeh2018multimodal}, which is applied in the context of sentiment analysis and provides a numerical effectiveness metric calculated from the fusion parameters, indicating how modality interactions contribute to the final prediction. Additionally, scores illustrating modality contributions (for two different representations of each modality) are used for an emotion recognition task in \cite{liu2022group}. Furthermore, \citet{wu2022interpretablecapsule} propose the use of capsules to obtain estimates on modality dynamics indicative of their contribution to the final emotion recognition prediction.
        
        \textbf{Ablation}: Ablation-based approaches can also be efficient in generating meaningful explanations. Specifically, the overall significance score of each modality is estimated in a medical signal classification application in \cite{ellis2021novel}. Additionally, \citet{lin2019explainable} introduce a feature importance method (for obtaining significance scores at the sensor level) and an ablation approach (for estimating feature importance at the signal level) in a multimodal affect recognition setting in \cite{lin2019explainable}.
        
        \textbf{Clustering}: Clustering techniques are also shown suitable in MXAI applications. In \cite{kumar2021towards}, the embeddings of each NN layer are used to demonstrate that deep layers distinguish emotions better than shallow layers in a multimodal emotion recognition model, utilizing audio and text features.
        
        \subsubsection{Multimodal task and multimodal explanation (\textbf{MM})}

        MM approaches constitute the most complex type, in terms of the number of modalities in the input and output feature spaces, since they support multimodal information both for the primary prediction task and the generated explanation. In the following, different types of MM MXAI methods are discussed, which are grouped according to the (most common) combinations of modalities/representations in the generated explanations.

\begin{figure*} [t]
            \centering
            
             \begin{subfigure}[t]{0.7\textwidth}
                \centering
                \includegraphics[width=\linewidth]{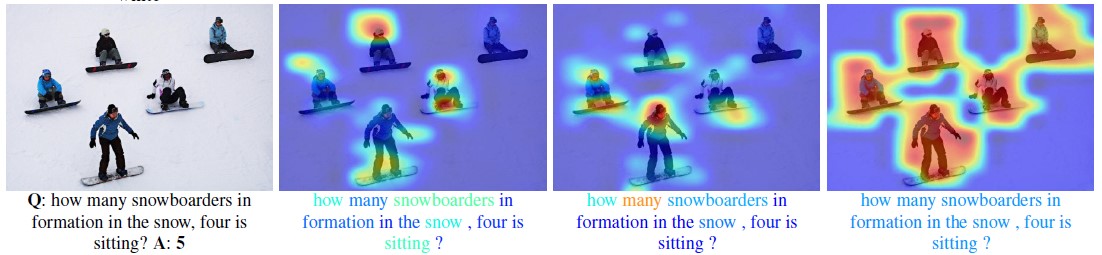} 
                \caption{} \label{fig:mm4}
            \end{subfigure}

            \begin{subfigure}[t]{0.3\textwidth}
            \centering
                \includegraphics[width=\linewidth]{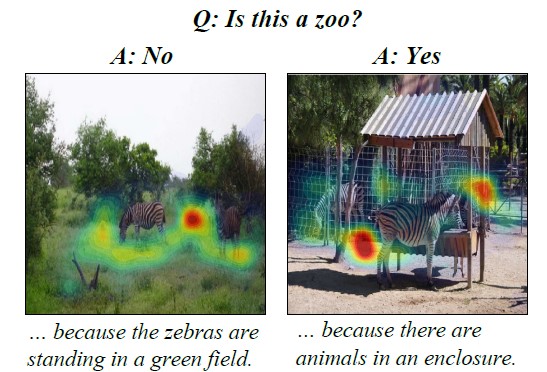} 
                \caption{} \label{fig:mm1}
            \end{subfigure}
            \hspace*{\fill}
            \begin{subfigure}[t]{0.25\textwidth}
            \centering
                \includegraphics[width=\linewidth]{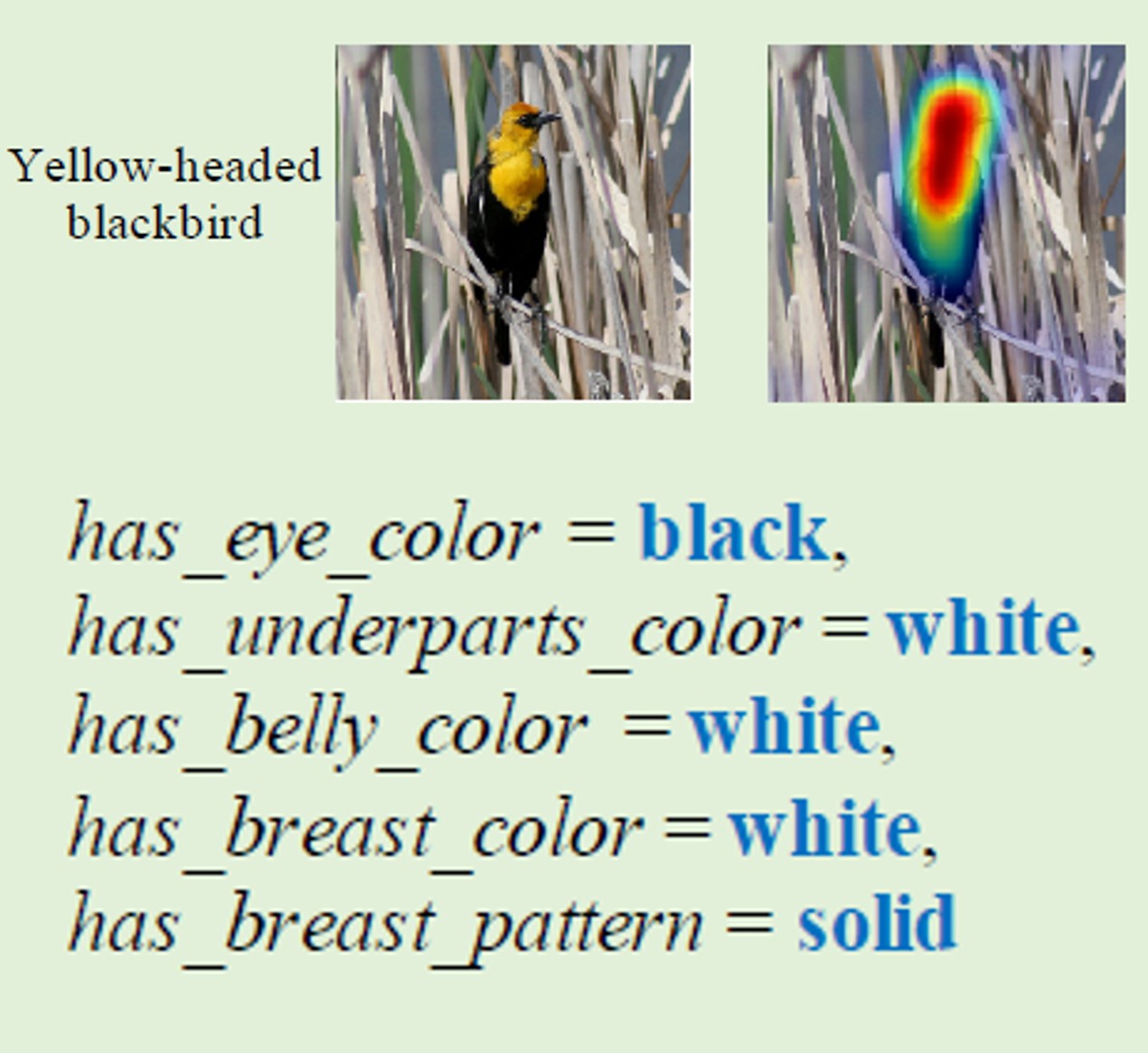} 
                \caption{} \label{fig:mm7}
            \end{subfigure}
            \hspace*{\fill}
            \begin{subfigure}[t]{0.35\textwidth}
                \centering
                \includegraphics[width=\linewidth]{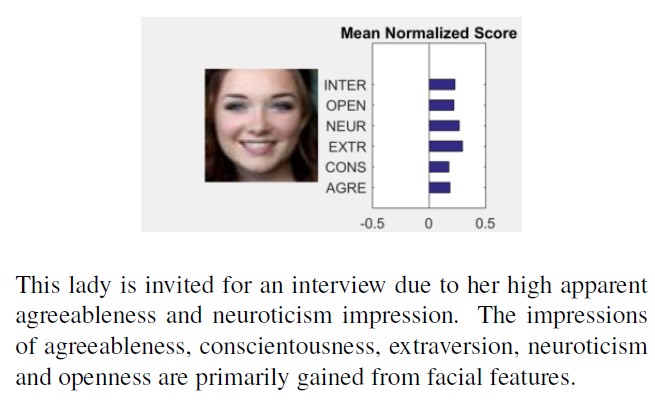} 
                \caption{} \label{fig:mm6}
            \end{subfigure}
        
            \caption{MM MXAI methods' generated explanations: a) Image- and text-heatmap \cite{Lu2016HiCoAt}, b) Image heatmap and text explanation \cite{Park2018}, c) Image heatmap and text attributes \cite{Selvaraju2018}, and d) Personality trait scores and text justification \cite{Kaya2017CVscreening}.}
            \label{fig:MMres}
        \end{figure*}

        \textbf{Image heatmap and text explanation}: The so far most popular MM MXAI category of methods relies on the use of a heatmap for the input image, along with supplementary text for generating a more complete explanation. In particular, an attention map and textual justifications are estimated for the VQA and the action recognition tasks in \cite{Park2018} (Fig. \ref{fig:mm1}). Similarly, \citet{li2018vqa} employ an attention mechanism for estimating important image regions and corresponding textual rationales are produced for complementing the explanation output for a VQA application. Interactive explanation schemes that allow users to interact with the model, when it provides erroneous answers, and subsequently to improve its behavior are employed in \cite{alipour2020study}, in order to investigate their efficiency in a VQA setting. Additionally, \citet{patro2020} adopt a correlation estimation approach between answers and explanations, in order to increase the robustness of the examined model, but also to provide meaningful explanations for the produced answer. Ground-truth visual and textual explanations are used in \cite{nagaraj-rao-etal-2021-first}, in order to train a model to generate corresponding multimodal explanations. Moreover, attention maps over video frames and respective textual justifications, regarding control predictions in a self-driving vehicle setting, are presented in \cite{Kim2020}.
        
        \textbf{Image- and text-heatmap explanation}: One of the most common techniques relies on the combination of heatmaps produced (separately) for both the image and the text modality. In particular, backpropagation and occlusion methods are used in \cite{Goyal2016}, in order to identify words (in questions) and image areas of significant importance when answering a visual question. Additionally, \citet{Lu2016HiCoAt} use attention schemes at multiple levels (between image and text phrases information streams) in a VQA application, in order to produce heatmaps over the input image and the respective question \cite{Lu2016HiCoAt} (Fig. \ref{fig:mm4}). Along with the textual answer in a VQA scenario, a justification is also produced in \cite{Wu2019}, based on the segmented image regions that correspond to specific words in the estimated rationale.

        \textbf{Image heatmap and text attributes}: Apart from combining image heatmaps with textual explanations, extracted text attributes can also be incorporated. In particular, domain-specific knowledge is employed in \cite{Selvaraju2018}, following an attribute-based formalism for the text stream and a Grad-CAM-grounded approach for the visual one in a zero-shot learning approach (Fig. \ref{fig:mm7}). Similarly, a gradient-based method for zero-shot learning and fine-grained classification is presented in \cite{Wickramanayake_Hsu_Lee_2019}, where, apart from the produced visual heatmap images, a text explanation generator is developed that takes into account the detected attributes (e.g. color, shape, etc.).
        
        \textbf{Other multimodal explanations}: The wide set of combinations that can be considered regarding the input modalities for the primary prediction task and the corresponding ones used for generating explanations allows for multiple and significantly diverse MM MXAI techniques to be developed. In particular, in an attempt to move towards cognition-level understanding, the so-called Recognition to Cognition Networks (R2C) are introduced in \cite{zellers2019recognition}, where, given a challenging question about an image, the networks target to answer correctly and then to provide a rationale justifying their answer; the explanations receive the form of bounding boxes that are associated with semantic concepts in the generated rationale. Additionally, \citet{cao2020behind} investigate the information learned in visual-language transformer models, aiming at identifying which is the most significant modality that captures the cross-modal interactions from certain attention heads, as well as the hidden visual or linguistic information that is stored in the latter. Moreover, an image with the aligned face of a candidate, normalized personality trait scores and a respective rationale are produced in \cite{Kaya2017CVscreening}, in order to explain the outcome of an interview in a job screening scenario (Fig. \ref{fig:mm6}).
        
    \subsection{MXAI Categorization Based on the Explanation Stage}  \label{MXAIcat2}
    
    MXAI methods can also be grouped into different categories based on the development/deployment stage (with respect to the primary prediction task model) at which explanations are learned/produced, as discussed in Section \ref{MXAIdefinitions} and detailed in the remainder of this section. 
        
        \subsubsection{Intrinsic}

        The methods of this category make use of the primary prediction model’s parameters to estimate meaningful interpretations of the produced results. In the following, intrinsic MXAI approaches are discussed in detail, while they are organized based on the most common types of generated explanations.

        \begin{itemize}
        \item \textbf{Attention heatmap}: Importance masks are efficient for demonstrating the components of the input data that are critical for explaining the model’s inference process. In particular, a so-called Multimodal Compact Bilinear (MCB) pooling operator is introduced in \cite{MCB2016}, in order to address the problem of intractable outer product calculation between matrices in VQA settings; the operator is used twice for creating spatial attention maps over the image to justify the estimated answer. \citet{yang2016stacked} introduce the so-called Stacked Attention Networks (SANs) (Fig. \ref{fig:I3}), which follow a multi-step reasoning approach; each layer identifies individual parts of the input image that the SAN has attended to produce answers for a VQA task. Additionally, a stochastic and a deterministic approach are presented in \cite{xu2015show}, in order to compute the attention weights for an image captioning model and to estimate the corresponding importance maps for each generated word. In order to force the model to focus on the same image regions that a human would do, an example-based approach is followed in \cite{patro2018differential}, where a nearest-neighbor method is adopted concerning the semantic similarity of images and distances between attention weights of similar examples are maintained to be lower than those of counter-examples. \citet{patro2019u} adopt a gradient-based approach for computing the loss gradients and attention maps are subsequently produced for the answers in a VQA setting, along with the corresponding uncertainties in the prediction process. Bidirectional Encoder Representations from Transformers (BERT) attention maps are investigated in a VQA application in \cite{alipour2020impact}, which are shown to concentrate on the most relevant areas in the examined image. Moreover, attention schemes are used to decompose the model’s reasoning into individual/consecutive steps that converge to the final generated answer in a VQA problem in \cite{mascharka2018transparency}. Furthermore, \citet{chen2019personalized} employ attention schemes in recommender systems, in order to detect important regions in product images that significantly affect the recommendation decision, taking into account user/item properties and the corresponding image. Text, tabular and graph data are considered by a self-attention mechanism in \cite{vijayaraghavan2021interpretable}, aiming at estimating salient features in hate-speech detection.

\begin{figure*} [t]
\centering
\begin{tabular}{cc}
\begin{tabular}{c}\begin{subfigure}[t]{0.45\textwidth}\includegraphics[width=\linewidth]{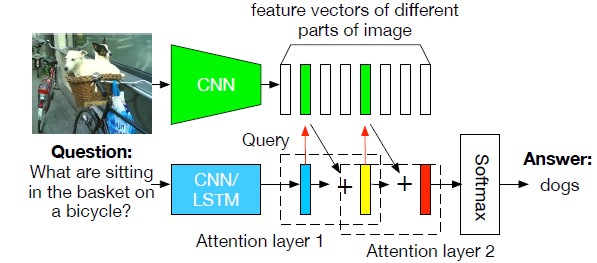}\caption{}\label{fig:I3}\end{subfigure}\end{tabular}&
\begin{tabular}{c}\begin{subfigure}[t]{0.45\textwidth}\includegraphics[width=\linewidth]{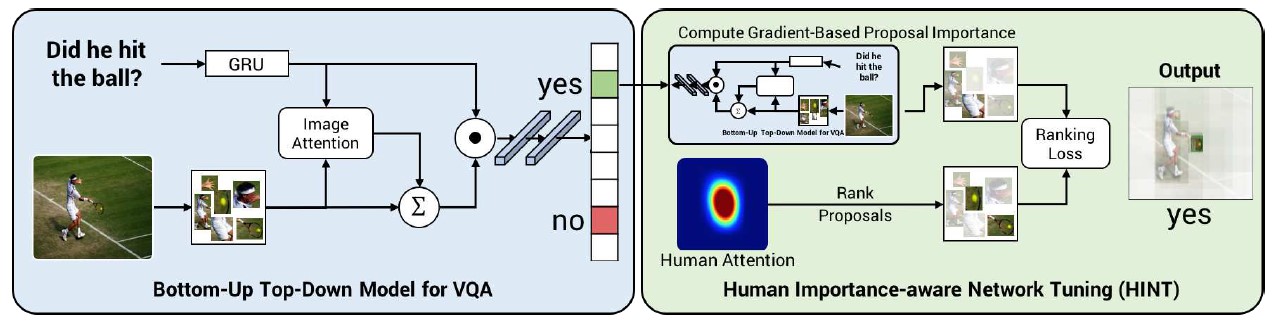}\caption{}\label{fig:I2}\end{subfigure}\end{tabular}\\
\end{tabular}
\begin{tabular}{c}
\begin{subfigure}[t]{0.65\textwidth}\includegraphics[width=\linewidth]{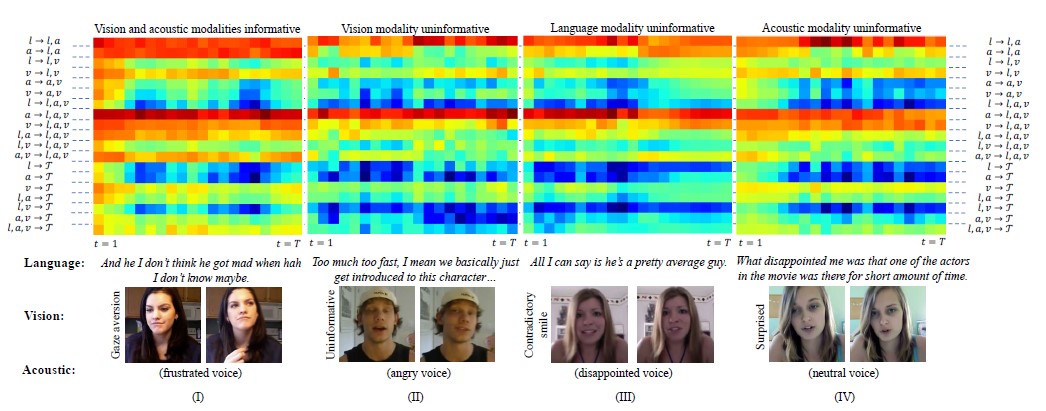}\caption{}\label{fig:I5}\end{subfigure}
\end{tabular}\\
\caption{Intrinsic MXAI methods' generated explanations: a) Attention heatmap \cite{yang2016stacked}, b) Image and attention heatmap \cite{selvaraju2019taking}, and c) Modality importance \cite{zadeh2018multimodal}.}
\label{fig:IntrArch}
\end{figure*}
        
        \item \textbf{Image and attention heatmap}:  In order to estimate more complete and informative explanations, attention heatmaps have also been combined with parts of the input image. In particular, an LSTM model is introduced in the context of a VQA task in \cite{zhu2016visual7w}, where a spatial attention mechanism relates words in the provided questions with corresponding image regions and eventually estimates bounding boxes of critical detected objects (over an attention heatmap) to ground the produced answer. Similarly, attention weight parameters are considered in \cite{anderson2018bottom}, in order to estimate salient regions that relate to each word in the generated captions and answers in a VQA task. Additionally, \citet{zhang2019interpretable} make use of an attention-based model within a supervised setting in a VQA task, in order to estimate region/object-based groundings in the examined image. \citet{selvaraju2019taking} follow a Grad-CAM-oriented approach for obtaining salient image regions and associating them with saliency scores (Fig. \ref{fig:I2}). Moreover, a self-attention mechanism is used for both images and text in a visual entailment task in \cite{xie2019visual}, in order to generate either heatmap- or image-based explanations.
        
        \item \textbf{Modality importance}: A critical aspect in MXAI analysis comprises the assessment of the level of importance of the various modalities involved in the primary prediction task. In this respect, a graph-based fusion approach is implemented in \cite{zadeh2018multimodal}, in order to dynamically model the interactions between modalities in the emotion recognition and sentiment analysis tasks (Fig. \ref{fig:I5}). Additionally, \citet{tsai-etal-2020-multimodal} estimate local and global explanations for emotion recognition, where dynamically adjusted importance measures are assigned to unimodal and bimodal interactions for each processed data sample. Moreover, a so-called capsule network is integrated into a routing mechanism in \cite{wu2022interpretablecapsule}, in order to estimate a contribution score for each modality involved in an emotion recognition task. An attention-based scheme is also used in \cite{liu2022group} to align text and speech representations in an emotion recognition setting; eventually, the extent of contribution of each modality to the final outcome is calculated. Furthermore, \citet{cao2020behind} make use of the learned attention weights and heads of visual-language models, in order to visualize what type of information has been encoded in each head, which modality is more important and the interactions of the latter.
        \end{itemize}
        
        \subsubsection{Post-hoc}
        
        MXAI approaches under this category aim to produce meaningful explanations of a model’s behavior after the AI prediction module has been applied and its results are made available, i.e. the model for the primary prediction task is considered as a ‘black box’ one. In the following, post-hoc MXAI approaches are presented in detail, while they are grouped taking into account the relevant (and most popular) primary prediction tasks.
            
        \begin{itemize}
            \item \textbf{Visual classification and captioning}: An attention mechanism is employed to provide saliency maps (both spatial and temporal) for each generated word for the tasks of image and video captioning in \cite{ramanishka2017top}; the mechanism takes into account the decrease observed in the output word probabilities when frames or image regions are removed from the input (Fig. \ref{fig:ph2}). Additionally, an inherently explainable decision tree is employed in \cite{Kaya2017CVscreening}, in order to interpret positive/negative outcomes, when deciding about interview invitations. A zero-shot learning approach for image-level classification is introduced in \cite{Selvaraju2018}, which targets the modeling of semantically meaningful concepts that have been implicitly learned by individual neurons in convolutional neural networks and produces a heatmap and the top activated attributes. Moreover, a back-propagation-based approach is adopted in \cite{xu2020model}, in order to identify the extent of the detected attributes’ (e.g. color, shape, etc.) contribution to the final outcome, while also producing complementary saliency maps and textual justifications (Fig. \ref{fig:ph1}). \citet{Gulshad2020} select examples from an opposing semantic class while considering visual attributes extracted by a model trained in an adversarial way; the explanation is eventually formed by combining attributes/examples of the original and the counter class. Furthermore, a graph representation is produced in \cite{aakur2018inherently}, incorporating concepts detected in the input video; then, an interactive agent is applied on top to provide explanations of the depicted action.
        
            \item \textbf{Visual question answering}: A guided backpropagation method is combined with a part occlusion one (applied to segments of input images and questions) in \cite{Goyal2016}, in order to identify visually important image regions and to predict the contribution of individual question words in the VQA generated answer (Fig. \ref{fig:ph3}). Additionally, \citet{ghosh2019generating} employ an attention-based heatmap to define the most relevant parts in an image, while a scene graph is also incorporated, along with NLP technologies, to provide natural language rationales, using the detected entities and their relations.

           \begin{figure*} [t]
                \centering
            
                \begin{subfigure}[t]{0.49\textwidth}
                \centering
                    \includegraphics[width=\linewidth]{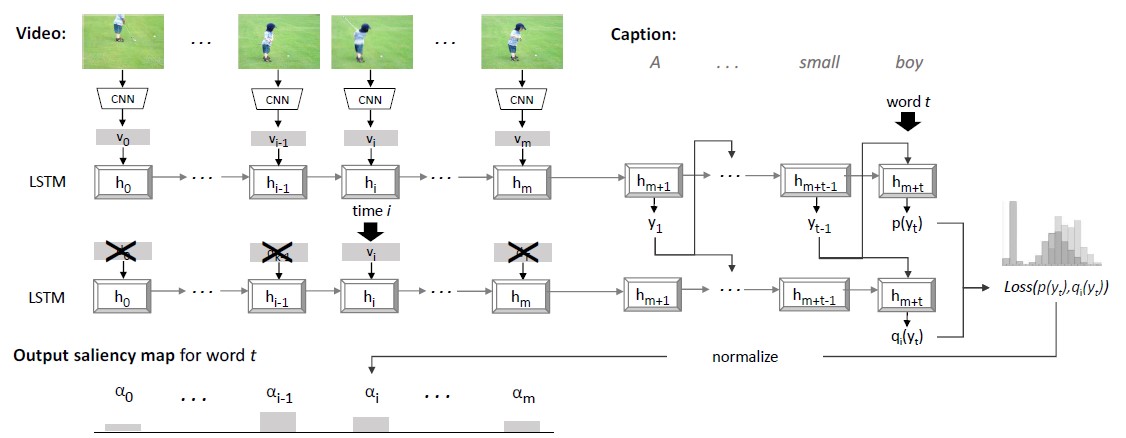} 
                    \caption{} \label{fig:ph2}
                \end{subfigure}
                \begin{subfigure}[t]{0.49\textwidth}
                \centering
                    \includegraphics[width=\linewidth]{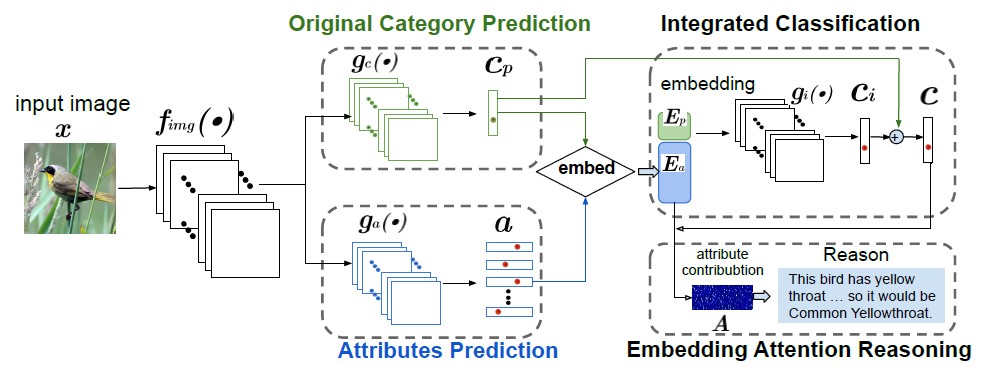} 
                    \caption{} \label{fig:ph1}
                \end{subfigure}

                \begin{subfigure}[t]{0.49\textwidth}
                \centering
                    \includegraphics[width=\linewidth]{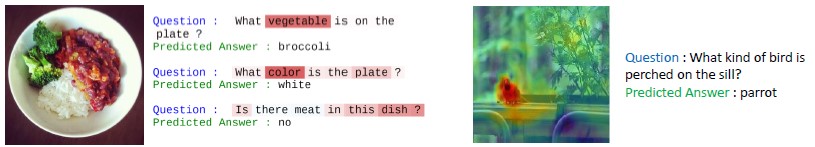} 
                    \caption{} \label{fig:ph3}
                \end{subfigure}
                \begin{subfigure}[t]{0.49\textwidth}
                \centering
                    \includegraphics[width=\linewidth]{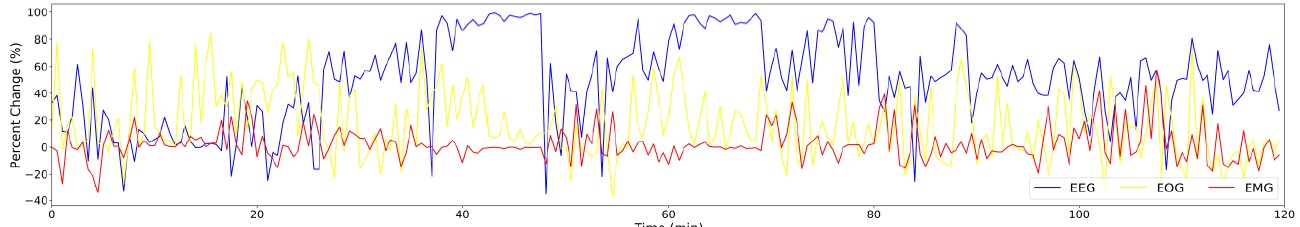} 
                    \caption{} \label{fig:ph4}
                \end{subfigure}
                
                \caption{Post-hoc MXAI methods' generated explanations: a) Video captioning \cite{ramanishka2017top}, b) Fine-grained visual classification \cite{xu2020model}, c) VQA \cite{Goyal2016}, and d) Biomedical signal processing \cite{ellis2021novel}.}
                \label{fig:PHArch}
            \end{figure*}
            
            \item \textbf{Biomedical signal processing}: A random forest algorithm is adapted to a multimodal affect recognition setting in \cite{lin2019explainable}, in order to provide sensor-level feature importance assessments and signal-level explanations. Additionally, principal components of layered embeddings of audio and text features are considered in \cite{kumar2021towards}, in order to reason about the separation of the detected emotion classes. \citet{ellis2021novel} apply an ablation-oriented analysis at the modality level, in order to determine the significance of different signals in sleep stage classification, by replacing one of the captured modalities with noisy data and observing the variation in the prediction probabilities (Fig. \ref{fig:ph4}).
        \end{itemize}

        \subsubsection{Separate module}
        
        This category comprises methods that rely on the development of a new/distinct module that generates explanations of the primary model’s behavior.

        \paragraph{\textbf{Joint training}} Under this consideration, the models for the primary prediction and the explanation generation tasks are constructed together, allowing in this way the training process of each module to affect the respective procedure of the other one. In the following, joint training MXAI approaches are discussed, while being grouped according to the primary task of concern.
        \begin{itemize}
            \item \textbf{Visual question answering}: An explanation module is trained together with the actual VQA model in \cite{goyal2017making}, in order to provide a similar image with a different answer (for a given question), essentially forming an example-based explanation. An attention-based mechanism (for the visual information stream) and an LSTM architecture (for the textual information stream) are combined on top of the actual model for producing the required explanations in \cite{Park2018}; the models can be trained either jointly or incrementally, depending on the selected task (namely VQA or action recognition, respectively). Additionally, a multi-task learning architecture is introduced in \cite{li2018vqa}, where a computational model is trained to generate an explanation, along with the answer predicted by the primary model (Fig. \ref{fig:jt5}). On the other hand, symbolic programs, which are used to represent the model’s reasoning, are utilized in \cite{vedantam2019probabilistic}, in order to justify the generated answers. \citet{patro2020} make use of a generator module (for producing textual explanations) and a respective correlation one (for ensuring that the produced answer complies with the estimated explanation). Moreover, an interactive approach is followed in \cite{alipour2020study}, where an attention-based scene graph, which takes into account the detected objects, is incorporated. Explanations are used in a competitive fashion to improve both the answers and the corresponding rationales in \cite{wu2020improving}. Furthermore, \citet{nagaraj-rao-etal-2021-first} make use of ground-truth visual and textual explanations for learning to interpret the generated answer. 
            
            \item \textbf{Visual classification}: An image classifier is jointly trained with a respective textual explanation generation module in \cite{Hendricks2016}, following a reinforcement learning approach. Similarly, a reasoning-based method is developed in \cite{barratt2017interpnet}, in order to rationalize classification decisions, using the activations of the fully connected layers of the classification model as input to the explanation generation one. Additionally, \citet{Hassan2019} introduce an LSTM-based explanation module equipped with two attention mechanisms (one applied to the considered attributes and one operating on top of the target categories), in order to produce rationales for the generated prediction and also to (supplementarily) retrieve semantically similar images. A saliency map and an associated attribute-based textual justification are used in \cite{Liu2020} (Fig. \ref{fig:jt1}), where the joint training strategy is shown to lead to significant performance improvements, compared to the separate training case. Moreover, \citet{zhuo2019explainable} exploit scene graph-based representations (making use of objects, their attributes and object relations information), aiming at both tracking the illustrated actions across the video frames and demonstrating the model’s reasoning process. 

        \begin{figure*} [t]
            \centering
            \begin{subfigure}[t]{0.49\textwidth}
            \centering
                \includegraphics[width=\linewidth]{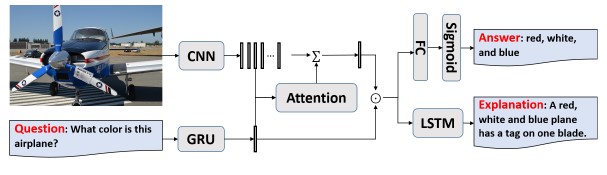} 
                \caption{} \label{fig:jt5}
            \end{subfigure}
            \hspace*{\fill}
            \begin{subfigure}[t]{0.49\textwidth}
            \centering
                \includegraphics[width=\linewidth]{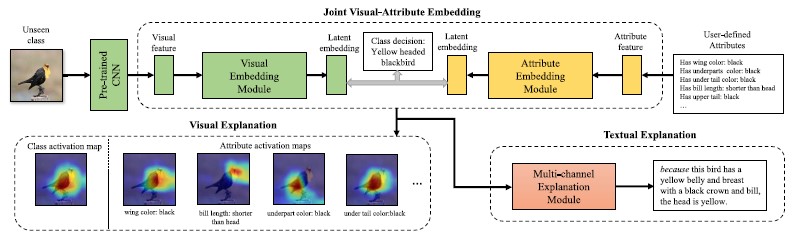} 
                \caption{} \label{fig:jt1}
            \end{subfigure}
            
            \begin{subfigure}[t]{0.49\textwidth}
            \centering
                \includegraphics[width=\linewidth]{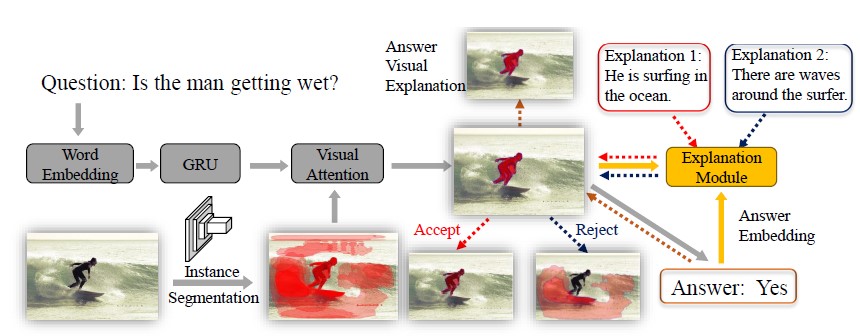} 
                \caption{} \label{fig:jt3}
            \end{subfigure}
            \hspace*{\fill}
            \begin{subfigure}[t]{0.49\textwidth}
            \centering
                \includegraphics[width=\linewidth]{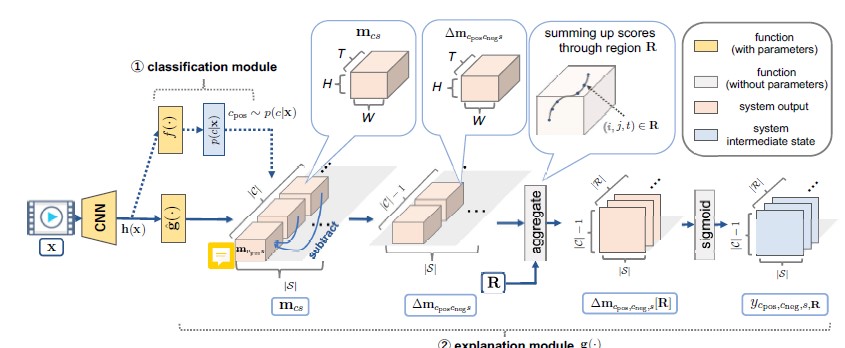} 
                \caption{} \label{fig:jt2}
            \end{subfigure}
            
            \caption{Separate module MXAI methods' generated explanations: a) Joint training for VQA \cite{li2018vqa}, b) Incremental and joint training for zero-shot learning \cite{Liu2020}, c) Incremental training for VQA \cite{Wu2019}, and d) Incremental training for video classification \cite{Kanehira2019}.}
            \label{fig:XstageArch}
        \end{figure*}
        
            \item \textbf{Visual captioning}: Topics extracted using LDA analysis are used in \cite{dong2017improving}, in order to jointly train a model with both a negative log-likelihood (for the primary captioning task) and an interpretable loss (for measuring the feature agreement with respect to semantically similar topics). Additionally, \citet{Han2018} design a module for examining whether the captioning model has focused on relevant image objects, by measuring the compliance of the detected objects/concepts and the correspondingly generated words using an attention mechanism.
            
            \item \textbf{Other}: \citet{lu2016visual} concentrate on detecting relationships among image objects and formulating a respective explanation graph, trained in an end-to-end fashion. Textual rationales are generated for answering follow-up questions regarding the examined image in a visual dialog setting in \cite{Das_2017_CVPR}. Additionally, responses to visual questions and textual rationales are jointly produced in the context of a visual common-sense reasoning task in \cite{zellers2019recognition}. \citet{Kim2020} introduce a combined approach for simultaneously realising next action prediction and justification in a self-driving cars application case. Moreover, recommendation and explanation functionalities are jointly developed in a reasoning framework in \cite{lin2019explainableoutfit}, aiming at providing explanations for each suggested item. 
        \end{itemize}
        
        \paragraph{\textbf{Incremental training}} In many cases, the explanation generation module is constructed separately from the respective primary task one (namely after the primary AI model is developed), due to factors like reduction in required training computations, lack of availability of explanations in the training data, etc. In the following, incremental training MXAI methods are presented that are grouped with respect to the primary task of concern.
        
        \begin{itemize}
            \item \textbf{Visual question answering}: \citet{Park2018} estimate both a textual rationale and a visual attention heatmap, by jointly or incrementally training the explanation module (depending on the task at hand). Additionally, \citet{li-etal-2018-tell} initially extract attributes (e.g. sit, phone, bench, etc.) and generate textual explanations for the image, while a reasoning module subsequently utilizes the extracted justifications to infer an answer to the given question. Attention mechanisms are used over segmented images (Fig. \ref{fig:jt3}) in \cite{Wu2019}, in order to produce explanations consistent with the predicted answers. A transformer-based model is employed for providing justifications in various vision-language tasks in \cite{marasovic-etal-2020-natural}, making use of pre-trained GPT-2 language models. 
            
            \item \textbf{Visual classification}: A so-called Deep Multimodal Explanation (DME) model, which incorporates a joint visual-attribute embedding module and a multi-channel explanation one trained in an end-to-end fashion, is introduced in \cite{Liu2020} (Fig. \ref{fig:jt1}), addressing the needs of explainable zero-shot learning applications. An explainable AI agent is developed in \cite{hendricks2018grounding}, where a phrase-critic model receives as input an image and a candidate explanation, and outputs a score indicating how good the candidate explanation is grounded on the image. Additionally, a method to generate textual counterfactual explanations is presented in \cite{hendricks2018counterfactuals}, focusing on inspecting which evidence data in the input is missing, but which could potentially contribute to a different classification decision if present in the examined image. \citet{Kanehira2019} exploit a spatio-temporal video region (tube) and textual attributes (e.g. using a pole, flipping, etc.) for estimating counterfactual multimodal explanations (Fig. \ref{fig:jt2}). Linguistic explanations and a set of visual examples for rendering the classification decision interpretable are used in \cite{kanehira2019learning}, where explanations are parameterized by three different NNs (namely a predictor, a linguistic explainer and an example selector model). Moreover, \citet{Wickramanayake_Hsu_Lee_2019} generate post-hoc linguistic justifications to rationalize the decision of a CNN, where a decision-relevance metric that measures the faithfulness of an explanation to a model’s reasoning is used. Furthermore, \citet{Lee2019CADmultimodal} introduce an explanation module that can be stacked to any Computer-Aided Diagnosis (CADx) network model, in order to provide rationales for medical diagnosis application cases.
            
            \item \textbf{Other}: \citet{Kim2018} produce introspective explanations in a self-driving vehicles scenario, which combine a visual (spatial) attention model (that identifies image regions that potentially influence the model’s output) and an attention-based video-to-text module (that produces textual explanations of the primary model's actions). Additionally, \citet{fang2019modularized} introduce textually-grounded explanations in computer vision applications, where the text descriptions are decomposed into three levels (namely entity, semantic attribute and color information) for progressively realizing compositional justification.
        \end{itemize}
        
    \subsection{MXAI Categorization Based on the Adopted Methodology}  \label{Methodologies}
 
    Within the field of MXAI research, there are certain commonly met methodologies (i.e. mathematical formalizations and mechanisms) that constitute fundamental building blocks of the respective proposed methods, as discussed in Section \ref{MXAIdefinitions}. To this end, using the particularly adopted methodology as a classification criterion, MXAI approaches can be categorized as follows:
    \begin{itemize}
        \item \textbf{Casual-modeling}: It mainly corresponds to techniques that support explainability through the use of counterfactual examples. In particular, counterfactual explanations aim at determining which are the minimum allowed modifications to the input data, so as to alter the primary model's prediction to a specific/pre-defined, yet different, output \cite{Kanehira2019, hendricks2018counterfactuals}.
        \item \textbf{Reasoning}: Similarly to numerous approaches in various AI-boosted application domains, extensions of the traditional multi-task learning conceptualization are also applied to the MXAI field. In particular, the examined AI model is jointly trained so as, apart from providing only predictions for the primary task, also to produce explanations about the generated outcomes and the overall model's reasoning process \cite{Park2018, Hendricks2016, Kim2018}.
        \item \textbf{Graph-modeling}: Analysis based on graphs constitutes a particularly valuable approach in MXAI scenarios, since they are highly efficient in identifying correlations among data points (also of diverse nature), while they also enable the generation of insightful representations/visualizations of the detected relations. In particular, graphical models have been extensively used in computer vision applications, where, for example, scene graphs have been employed for representing an image and for subsequently estimating answers/explanations for a given question \cite{alipour2020study}. Similarly, graphs have been utilized for estimating symbolic representations of textual sources, targeting, for example, to produce a step-by-step representation of the AI model's  inference process \cite{mascharka2018transparency}.
        \item \textbf{Attribute-based}: Identifying attributes in the input data, which are critical for shaping the model's predictions, constitutes an efficient way for subsequently producing accurate explanations of the model's behavior. For the case of visual input data, attributes can refer to certain image characteristics (e.g. color and shape of objects), aiming at eventually estimating an overall image saliency map \cite{Liu2020}. In a similar fashion, natural language rationales can be formed to explain individual model's decisions, e.g. determining the specific attributes/arguments behind a bird being classified as belonging to a particular breed \cite{Hassan2019}.
        \item \textbf{Interactive}: Approaches under this category allow humans to interactively intervene in the explanation generation process. Specifically, the user is allowed to provide feedback about the model's produced predictions and, subsequently, the model exploits the collected information for improving its performance or posing questions about the assessment of the results. For example, humans are capable of evaluating AI models' predictions and corresponding explanations through an interactive framework in \cite{alipour2020study}, while follow-up questions are presented to the user for justification purposes in \cite{Das_2017_CVPR}.
        \item \textbf{Fusion-based}: Numerous and diverse fusion schemes have been adopted in multimodal prediction models for estimating comprehensive data representations and achieving accurate prediction results. For example, a multiplication operator of image and question representations is typically used in VQA approaches. In this context, investigating the way that fusion is applied and the effects that the latter mechanism imposes can provide significant insights in terms of explainability purposes, e.g. identifying how each modality contributes to the produced outcome \cite{liu2022group}.
        \item \textbf{Attention-based}: The scope of the attention mechanism is to enable the trained model to focus only on specific features in the input data space that are important for realizing robust predictions, e.g. specific image patches or individual words in textual phrases. The latter is achieved by estimating weights that modulate the model’s attention on the input data. The estimated computed weights can be utilized though for providing meaningful explanations of the model’s decisions, e.g. in the form of visualizations of the  attended regions in an image or words in a sentence \cite{Park2018, Lu2016HiCoAt, anderson2018bottom, patro2019u}.
    \end{itemize}

\section{Evaluation of generated explanations} \label{evaluation} 
    
Evaluation schemes in XAI applications aim at assessing the accuracy and efficiency of the generated explanations, i.e. the ability to which the XAI method at hand explains the behavior/decisions of the model developed for the primary prediction task. The main categories to which XAI (including MXAI techniques) evaluation approaches can be roughly classified to comprise \cite{doshi2017towards}: a) Application-grounded, i.e. implementing experimental evaluations with domain experts, b) Human-grounded, i.e. conducting experiments with humans, without necessarily being domain experts in the examined field, and c) Functionally-grounded, i.e. using a formal definition of explainability in order to assess the quality of the produced explanations. With respect to the particular case of MXAI methods, the evaluation protocol can consider/assess the involved modalities independently (with textual and visual explanations being the most common ones) or also take into account cross-modal inter-relations/dependencies.

    \subsection{Evaluation of Textual Explanations}
    
    In order to assess the quality (i.e. accuracy, correctness, etc.) of the generated textual explanations (that are usually in the form of natural language sentences), a large body of the relevant literature relies on the implementation of user-centered studies, i.e. experimental procedures that require the provision of human-user assessments (e.g. in the form of questionnaires, explanation gradations/rankings, etc.) \cite{Park2018, Kim2018, Hendricks2016, hendricks2018grounding, wu2020improving, marasovic-etal-2020-natural}. The latter is mainly due to: a) Lack of publicly available datasets that include ground-truth explanation-related annotations, and b) Inherent difficulty in determining precisely and unequivocally what a ‘good’ explanation is for a given data input. However, when ground-truth explanation data are available, the following main metrics (where BLEU-1,2,3,4, ROUGE, METEOR, CIDEr and SPICE have been borrowed from the NLP literature) have been widely used, in order to evaluate the similarity between the generated textual justification and the corresponding target/ground-truth one:
    \begin{itemize}
    \item \textbf{BLEU}: It counts an n-gram (i.e. continuous sequence of n words) based matching score  between the candidate explanation and the ground-truth one, regardless of the word order \cite{Hassan2019, Liu2020, Park2018, patro2020, Wu2019, Lee2019CADmultimodal, wu2020improving, nagaraj-rao-etal-2021-first, li-etal-2018-tell, barratt2017interpnet, li2018vqa, Kim2020, lu2022learn}.
    \item \textbf{ROUGE}: Among the various individual metrics belonging to this category, ROUGE-L takes into account the longest common sub-sequence between the reference and the predicted explanation phrase \cite{Hassan2019, Liu2020, Park2018, patro2020, Wu2019, Lee2019CADmultimodal, wu2020improving, nagaraj-rao-etal-2021-first, li-etal-2018-tell, barratt2017interpnet, Kim2020, li2018vqa, lu2022learn}.
    \item \textbf{METEOR}: While originally designed to address shortcomings of the respective BLEU score, its estimation is based on an explicit word-to-word matching between the generated explanation and all respective reference ones \cite{Hassan2019, Liu2020, Park2018, patro2020, Kim2018, Hendricks2016, Wu2019, wu2020improving, nagaraj-rao-etal-2021-first, li-etal-2018-tell, barratt2017interpnet, Kim2020, li2018vqa}.
    \item \textbf{CIDEr}: It assesses the degree of consensus between a candidate explanation sentence and a set of reference ones, by examining how often n-grams in the candidate phrase appear in the reference ones \cite{Hassan2019, Liu2020, Park2018, patro2020, Kim2018, Hendricks2016, Wu2019, Lee2019CADmultimodal, wu2020improving, nagaraj-rao-etal-2021-first, li-etal-2018-tell, barratt2017interpnet, Kim2020, li2018vqa}.
    \item \textbf{SPICE}: It is based on the creation of semantic scene graphs from dependency parse trees created using the candidate and reference explanation sentences, taking into account objects, attributes and their relationships; it relies on assessing the generated explanation/sentence quality using an F-score metric calculated over tuples (conjunction of logical propositions) belonging to both graphs \cite{Park2018, patro2020, Wu2019, wu2020improving, Kim2020}.
    \item \textbf{Ratio of unique or novel sentences}: These focus on identifying/counting the explanation sentences that have not been generated before and those that do not exist in the training set \cite{wang2017skeleton}.
    \item \textbf{Cosine-similarity}: It measures the correspondence of the generated natural language explanations and the ground-truth ones, taking into account the respective available semantic embeddings \cite{reimers2019sentence}.
    \item \textbf{Phrase error and accuracy with counterfactual test}: These metrics are applicable to the case of counterfactual textual explanations \cite{hendricks2018counterfactuals}. In particular, phrase error estimates the degree of similarity of the counterfactual and the ground-truth sentence, i.e. an ideal score should be equal to zero. Additionally, accuracy aims at measuring the decrease in class prediction performance, where additional counterfactual text is provided (along with the originally generated textual explanation) as input to the classifier.
    \item \textbf{Class similarity}: It relies on estimating the relevance of the generated explanation for a particular semantic class with the ground-truth ones for the same class \cite{Hendricks2016}, by employing the CIDEr metric between the generated explanations and all reference sentences of the examined class (and not just the ground-truth ones for the specific image in question).
    \item \textbf{Class ranking}: It is an extension of the class similarity concept to further validate whether the generated explanations are class relevant. It relies on the estimation of the similarity score for each examined sentence and every considered class \cite{Hendricks2016}. 
    \item \textbf{Relevance of visual attributes}: Such metrics measure the proportion of accurately represented ground truth attributes in the top-k generated explanations for each instance \cite{Selvaraju2018} or use the estimated attributes to assess the relevance of the generated explanations with the visual features that the examined model has learned \cite{Wickramanayake_Hsu_Lee_2019}. The latter is accomplished by counting the words that are common in the ground-truth description of an object and the generated justification, and then calculating the ratio of this count to the total number of words in the ground-truth description.
    \item \textbf{Relevance of textual attributes}: It evaluates the accuracy of the generated explanations (i.e. predicted attributes (in text form) and estimated captions), by utilizing a cosine similarity-based formalism. 
    \item \textbf{Word importance}: It is based on estimating the importance of the individual words comprising the generated explanations under different evaluation scenarios, e.g. removing the top-3 relevant words \cite{sun2020understanding}, consideration of part-of-speech tags \cite{Goyal2016}, etc.
    \item \textbf{Position of first and number of relevant explanations}: These estimate the ranking position of the first relevant explanation in the set of all generated ones and the number of relevant explanations in the top-5 produced ones \cite{ghosh2019generating}.
    \end{itemize}
        
    \subsection{Evaluation of Visual Explanations}
    Similarly to the case of evaluating textual justifications, a large portion of the MXAI literature that produces visual explanations relies on the implementation of user-centered studies. For the latter case, qualitative assessment of the produced outcomes is usually realized by superimposing the explanation (e.g. heatmap) to the input visual data (either image or video) and, subsequently, the human-user is involved, so as to judge whether the identified/highlighted areas correspond to truly important pixels/points for the primary model’s prediction. On the other hand, when relevant ground-truth information is available, quantitative evaluation can be performed by: a) Estimating the correspondence of the generated explanations (e.g. in the form of target bounding-boxes, human attention and segmentation maps, etc.) to the ground-truth ones, and b) Examining the contribution of individual image pixels and regions to the model decisions, by observing the effect of removing or superimposing them from/to the model's input data. When ground-truth explanation-related information is provided, the following metrics/approaches are commonly used in the literature:
    \begin{itemize}
    \item \textbf{Intersection-over-Union (IoU)}: This measures the correspondence between a given reference mask or bounding-box and the respective one associated with the explanation \cite{Liu2020, nagaraj-rao-etal-2021-first}. 
    \item \textbf{Mask alignment}: It evaluates the visual alignment between the ground-truth and the generated explanation masks, focusing on e.g. examining whether the maximum attention weights reside inside the ground-truth bounding-box area \cite{zhu2016visual7w}.
    \item \textbf{Average fraction of activations}: It estimates the average percentage of the points (belonging to the produced activation maps) that lie within the provided ground-truth bounding boxes \cite{Selvaraju2017}.
    \item \textbf{Earth Mover’s Distance (EMD)}: It measures the similarity/distance between the produced attention/explanation map and the human-provided one, while considering them as two different probability distributions \cite{Park2018, Wu2019, patro2019u}.
    \item \textbf{Rank correlation}: It relies initially on the ranking of the pixels of the produced/available heatmaps according to their spatial attention values and, subsequently, estimating the correlation between these ordered lists of pixels \cite{Das2017HumanAttention, Park2018, patro2020, Goyal2016, selvaraju2019taking, patro2018differential, patro2019u}.
    \item \textbf{Foreground Attention Rate (FAR)}: It measures the degree of compliance of the detected foreground objects with the provided attention heatmap over the examined image \cite{xu2020model}.
    \item \textbf{Quality of attention/saliency map}: It assesses the accuracy of the generated heatmaps, by estimating the degree of consistency with the respective human-annotated ground-truth ones \cite{zhang2018top}. 
    \item \textbf{Precision and recall}: The conventional precision and recall performance metrics are extended to the case of evaluating the generated visual explanations, by e.g. estimating how often the center of the generated attention map overlaps with the available ground-truth region annotation \cite{mascharka2018transparency}.
    \item \textbf{Attention correctness} \cite{liu2017attention}: Specifically designed for attention models used in visual captioning tasks \cite{sun2020understanding, ramanishka2017top}, it is equal to the normalized sum of all attention weights that relate to a given word that, at the same time, correspond to the respective ground-truth annotated regions.
    \item \textbf{Negative class accuracy}: It aims at assessing the validity of the produced counterfactual explanation, by examining the extent to which the output explanation changes when the estimated attended region is removed \cite{Kanehira2019}. 
    \end{itemize}

    Assessing the explanation evaluation issue from a causal relation modeling perspective, the analysis focuses on examining whether the generated visual regions contribute to the correct prediction and how changes in the input affect the final prediction result, i.e. not being constrained only to the case of examining solely the alignment of visual explanations with ground-truth data. In this context, the following measures/approaches are widely used:
    \begin{itemize}
    \item \textbf{Deletion}: It identifies critical pixels, which, when removed from an image, lead to a drop in detection performance for a given semantic class \cite{Petsiuk2018RISERI}.
    \item \textbf{Insertion}: It determines important pixels, which, when added to an image, lead to an increase in the detection performance for the examined semantic class \cite{Petsiuk2018RISERI}.
    \item \textbf{Average drop}: It corresponds to the average decrease in the model’s prediction performance for a given class, when only explanation-related information is provided as input to the model, instead of the full original image \cite{Chattopadhay2018}.
    \item \textbf{Average drop in deletion}: It is similar to the ‘average drop’ metric described above, but it makes use of the prediction score for the class of interest, when considering the inverted explanation heatmap \cite{jung2021towards}.
    \item \textbf{Increase in confidence}: It corresponds to the increase in the model’s prediction confidence, when providing only explanation-related information as input to the model, instead of the original image \cite{Chattopadhay2018}.
    \item \textbf{Win percentage}: It compares the performance of different methods, by estimating the frequency that the performance achieved by a given method, when using only the  generated explanation heatmap, is higher than that obtained by other/competitive approaches \cite{Chattopadhay2018}.  
    \end{itemize}

    \subsection{Multimodal Evaluation}

    While the wide majority of the literature approaches considers individual unimodal metrics for evaluating the quality of the generated MXAI explanations, methods that implement multimodal evaluation schemes (i.e. assessment approaches that take into account the correlations among the examined modalities) have also been proposed, as follows:
    \begin{itemize}
    \item \textbf{Concept accuracy} \cite{Kanehira2019}: This aims at evaluating multimodal (visual and textual) counterfactual explanations for video classification. The metric estimates the compatibility of the estimated words (i.e. assigned attributes, e.g. using pole, flipping, etc.) in the textual justification and their visual counterparts, by comparing the conventional IoU metric of a given attribute-bounding-box pair with the respective pairs corresponding to the remaining generated counterfactual explanations. 
    \item \textbf{Complementarity} \cite{kanehira2019learning}: It estimates the degree of complementarity between the visual and the textual part of an explanation, by utilizing the ``reasoner"'s (it receives as input the generated explanation pairs) score for the class identified by the ``predictor" (the actual prediction module) for each candidate combination of text and image explanation pairs.
    \item \textbf{Fidelity} \cite{kanehira2019learning}: This metric aims at examining how class predictions are obtained, by utilizing the generated explanations. The ``reasoner" (it receives as input the generated explanation pairs) and the ``predictor" (the actual prediction module) are compared in terms of accuracy and prediction consistency. 
    \end{itemize}

\section{Current challenges and future research directions} \label{challenges}  

Despite the large body of MXAI works that have recently been introduced, significant challenges and open research problems are still present, which if sufficiently addressed will further increase the efficiency and acceptance of MXAI schemes. It needs to be mentioned that the research directions described below are often applicable also to the unimodal XAI field.

\textbf{Convergence to formal and widely accepted definitions/terminology}. Although many research studies have recently appeared in the MXAI field, little to no formality has been adopted, concerning the definitions and terminology used. In particular, many researchers make use of ad-hoc descriptions to delineate their research activities, while they often define ‘explainability’ and ‘interpretability’ in various (and often conflicting) ways. As a result, no concrete and widely accepted terminology is currently present. Defining what an explanation is and how its accuracy/efficiency can be measured using well-defined qualitative/quantitative norms and experimental frameworks, apart from enhancing formalization aspects in the field, will also significantly facilitate the comparative evaluation of the numerous proposed explainability methods \cite{saeed2023XAIchallenges}. The latter will also greatly assist in addressing current controversies, like assigning different terms to similar methods or associating similar names with fundamentally different (algorithmic) concepts.

\textbf{Usage of attention mechanisms in explanation schemes}. Attention mechanisms, apart from being used in numerous data analysis tasks, have also been utilized for generating explanations of the prediction models' behavior, typically in the form of visualization schemes (indicating words or image areas where the primary model focuses on) or feature importance metrics. However, several concerns and conflicting arguments have emerged, raising fundamental doubts regarding the suitability of attention mechanisms to produce actual explanations. In particular, experimental studies show that distributions between learned attention weights and gradient-based feature relevance methods are not highly correlated for similar predictions \cite{Jain2019}; hence, conventional attention-based explanations can not be considered equivalent to others. However, contradictory experimental results move in the opposite direction, i.e. the usage of attention schemes for explanation generation is not always applicable, but it depends on the actual definition of explanation that is adopted in the particular application at hand \cite{wiegreffe2020attention}. In this context, more detailed and in-depth studies need to be conducted, in order to shed more light on whether and under which exact conditions attention schemes can be used for providing meaningful explanations, as well as how such methods relate to other non attention-based MXAI approaches.

\textbf{Generalization ability of MXAI methods}. The wide majority of the available methods has only been designed for specific AI model architectures, regarding the primary prediction task. For example, there is a significant number of methods that have been designed for the VQA scenario; however, such approaches have not been evaluated in other vision-language applications. Naturally, it can be well admitted that introducing model-specific explanation schemes is very restrictive and expensive. Robustly extending existing methods to other tasks and architectures would significantly save research resources and would likely lead to performance improvements.

\textbf{Extension of MXAI schemes to more than two explanation modalities}. Most MXAI methods focus on producing unimodal or bimodal explanations. However, extending these representations to higher dimensionality multimodal feature spaces (i.e. feature spaces that are composed of more than two modalities) would inevitably further increase the expressiveness and accuracy of the produced explanations.

\textbf{Estimation of causal explanations}. So far no significant attention has been given to the causality perspective of explanations, while causal relationships are the particular type of relations that humans inherently perceive. In this respect, causal explanations can enable the interpretation of how one event can lead to another one and, hence, the development of a deeper understanding of the world. On the other hand, identifying the factors that cause an event to occur can also facilitate the prediction of how the event might unfold and/or how it could/should be treated in the future \cite{saeed2023XAIchallenges}. Therefore, apart from identifying which features are important for a given model, how predictions are affected by modifications in the feature values is important to understand the model’s reasoning process itself. In the context of the multimodal setting, causality needs to be examined in terms of how each individual modality and the correspondingly particular features affect the prediction outcome (and not simply identifying which features are important).

\textbf{Removing bias in textual explanations}. The main paradigm being followed for estimating textual explanations consists of collecting natural language rationales from humans and subsequently developing/training an explanation module with these descriptions as ground truth data. However, human textual annotations (especially when it comes to long textual justifications) typically contain (contradictory) biases that are related to the particular background and temperament of each involved individual. To this end, developing modulation schemes for identifying/removing bias and resolving conflicting annotation cases would significantly improve the quality of the generated textual annotations.

\textbf{Lack of ground-truth visual explanations}. Contrary to the case of textual explanations, a lack of corresponding ground-truth visual explanation data, which would enable the development of explanation generation modules trained under a supervised learning scenario, is observed. Despite the fact that collecting such manual annotations can be very expensive, the availability of such data would greatly boost the development of more accurate and robust MXAI schemes.

\textbf{Insufficient MXAI evaluation}. One of the main gaps in MXAI research concerns the lack of standard evaluation metrics, protocols and benchmarks for assessing the quality of the produced explanations. This naturally hinders the objective comparative evaluation of different methods and the identification of the most efficient practices for generating accurate explanations. Additionally, the most widely adopted norm consists of examining each modality separately (i.e. not investigating the correlations among the different information streams of the produced explanations); hence, often leading to inaccurate observations and possible misconceptions regarding the obtained results. Moreover, the inherent human subjectivity, when it comes to explanation assessment, adds to the problem difficulty. Addressing all the aforementioned challenges, towards achieving objective MXAI evaluation results, should be coupled with the definition of suitable multimodal interpretability evaluation metrics, where only very few and highly task/model-specific ones have been introduced so far \cite{Kanehira2019, kanehira2019learning}.

\textbf{Explanations targeting specific domains and end-users}. A critical aspect of explainability is that of producing explanations that are tailored to specific end-users, considering their individual needs and diverse backgrounds. Interpretation capabilities should be inherently tied to the nature of the particular user's experience and expertise (something that is usually neglected), and in the majority of the scenarios the produced explanations cannot be well understood by non-experts. The design of explanation schemes should consider the diverse backgrounds, knowledge and cognitive capabilities of the end-user interacting with a particular AI system. Users with limited technical expertise may require conceptually simpler explanations that would support a clear understanding of the system's decision-making process. On the other hand, users with domain-specific expertise may require more detailed and accurate information. On the other hand, tailoring explanations to different user profiles would further enhance the user's experience, by fostering comprehension, building trust and enabling effective collaboration between humans and AI systems.

\section{Conclusions} \label{conclusion}     

In this paper, a comprehensive, systematic and in depth study regarding the developments in field of Multimodal eXplainable Artificial Intelligence (MXAI) was presented. Initially, an extensive analysis of the relevant primary prediction tasks (e.g. image/video captioning, visual question answering, etc.) and the corresponding publicly available datasets, where MXAI approaches have been applied so far, was provided. Subsequently, a thorough and structured presentation of the MXAI methods of the literature was given, based on the following key criteria: a) The number of the involved modalities (regarding both the primary prediction model input and the generated explanation feature spaces), b) The development/deployment stage (with respect to the primary prediction task model) at which explanations are learned/produced, and c) The type of the methodology (i.e. mathematical formalism/mechanism) adopted for producing the actual explanations. Then, a detailed discussion regarding the issue of MXAI methods' evaluation was provided, emphasizing on outlining the relevant quantitative performance metrics. Finally, a comprehensive analysis of current challenges and future research directions in the field was given.

\bmhead{Acknowledgements}
The research leading to the results of this paper has received funding from the European Union’s Horizon Europe research and development programme under grant agreement No. 101073924 (ONELAB).

\section*{Declarations}
\textbf{Funding} The research leading to these results received funding from the European Commission under Grant Agreement 101073924 (ONELAB).\\\\
\textbf{Conflict of interest} The authors have no competing interests to declare that are relevant to the content of this article.\\\\
\textbf{Author contribution} Nikolaos Rodis, Christos Sardianos and Georgios Th. Papadopoulos performed the literature review and prepared a draft of the manuscript. Panagiotis Radoglou-Grammatikis, Panagiotis Sarigiannidis and Iraklis Varlamis performed reviewing and editing of the manuscript. Georgios Th. Papadopoulos was responsible for receiving the funding to implement the study.

\bibliography{references}

\end{document}